\documentclass[runningheads]{llncs}
\usepackage{graphicx}
\usepackage{comment}
\usepackage{amsmath,amssymb} 
\usepackage{color}
\usepackage{cite}

\usepackage{caption}
\usepackage{subcaption}
\usepackage{tabularx}
\usepackage{multirow}
\usepackage{graphbox}
\usepackage[export]{adjustbox}

\usepackage[width=122mm,left=12mm,paperwidth=146mm,height=193mm,top=12mm,paperheight=217mm]{geometry}

\begin{document}
\pagestyle{headings}
\mainmatter
\def\ECCVSubNumber{1637}  

\title{Fast Bi-layer Neural Synthesis of One-Shot Realistic Head Avatars} 

\titlerunning{Fast Bi-layer Head Avatars}
%
\author{Egor~Zakharov\inst{1,2} \and
Aleksei~Ivakhnenko\inst{1} \and
Aliaksandra~Shysheya\inst{1,3} \and
Victor~Lempitsky\inst{1,2}}
\authorrunning{E. Zakharov et al.}
%
\institute{Samsung AI Center -- Moscow, Russia \and
Skolkovo Institute of Science and Technology, Russia \and
University of Cambridge, UK}

\maketitle

\begin{abstract}
We propose a neural rendering-based system that creates head avatars from a single photograph. Our approach models a person's appearance by decomposing it into two layers. The first layer is a pose-dependent coarse image that is synthesized by a small neural network. The second layer is defined by a pose-independent texture image that contains high-frequency details. The texture image is generated offline, warped and added to the coarse image to ensure a high effective resolution of synthesized head views. We compare our system to analogous state-of-the-art systems in terms of visual quality and speed. The experiments show significant inference speedup over previous neural head avatar models for a given visual quality. We also report on a real-time smartphone-based implementation of our system.
\end{abstract}

\newlength{\wid}

\section{Introduction}

\begin{figure}[t]
    \centering
    \includegraphics[width=\textwidth]{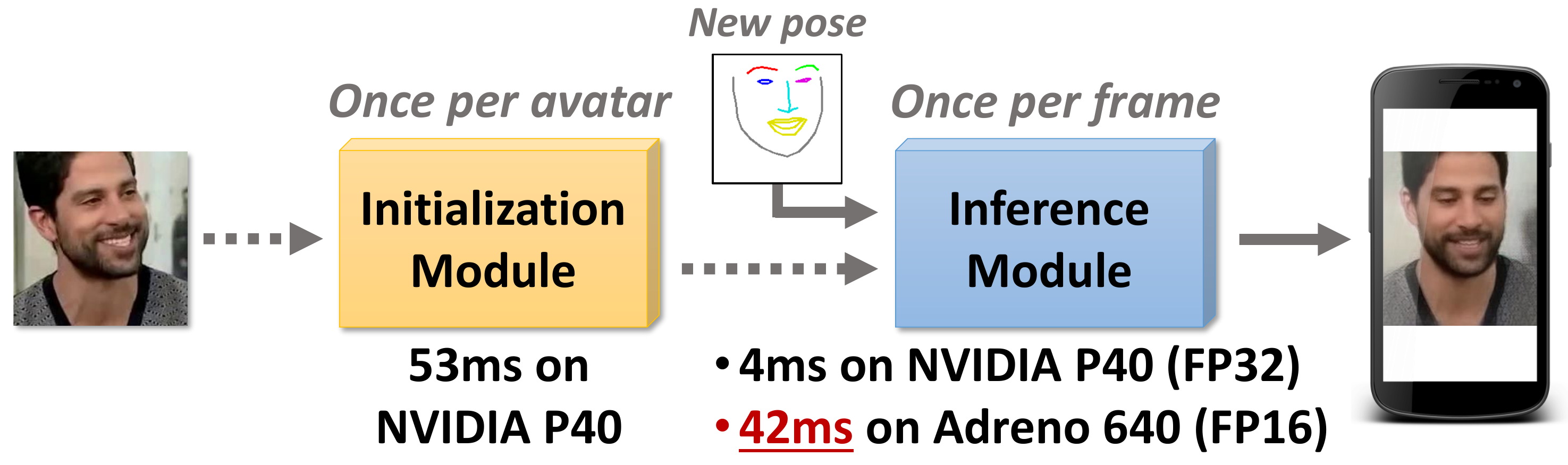}
    \caption{Our new architecture creates photorealistic neural avatars in one-shot mode and achieves considerable speed-up over previous approaches. Rendering takes just 42 milliseconds on Adreno 640 (Snapdragon 855) GPU, FP16 mode.}
\end{figure}

Personalized head avatars driven by keypoints or other mimics/pose representation is a technology with manifold applications in telepresence, gaming, AR/VR applications, and special effects industry. Modeling human head appearance is a daunting task, due to complex geometric and photometric properties of human heads including hair, mouth cavity and surrounding clothing. For at least two decades, creating head avatars (talking head models) was done with computer graphics tools using mesh-based surface models and texture maps. The resulting systems fall into two groups. Some~\cite{Alexander10} are able to model specific people with very high realism after significant acquisition and design efforts are spent on those particular people. Others~\cite{Hu17} are able to create talking head models from as little as a single photograph, but do not aim to achieve photorealism.

In recent years, \textit{neural talking heads} have emerged as an alternative to classic computer graphics pipeline striving to achieve both high realism and ease of acquisition. The first works required a video \cite{Kim18,Wang18c} or even multiple videos~\cite{Suwajanakorn17,Lombardi18} to create a neural network that can synthesize talking head view of a person. Most recently, several works \cite{Zakharov19,Tripathy19,Fu19,Siarohin19,Ha19,Siarohin19,Wang19} presented systems that create neural head avatars from a handful of photographs (\textit{few-shot} setting) or a single photograph (\textit{one-shot} setting), causing both excitement and concerns about potential misuse of such technology. 

Existing few-shot neural head avatar systems achieve remarkable results. Yet, unlike some of the graphics-based avatars, the neural systems are too slow to be deployed on mobile devices and require a high-end desktop GPU to run in real-time. We note that most application scenarios of neural avatars, especially those related to telepresence, would benefit highly from the capability to run in real-time on a mobile device. While in theory neural architectures within state-of-the-art approaches can be scaled down in order to run faster, we show that such scaling down results in a very unfavourable speed-realism tradeoff.

In this work, we address the speed limitataions of one-shot neural head avatar systems, and develop an approach that can run much faster than previous models. To achieve this, we adopt a \textit{bi-layer representation}, where the image of an avatar in a new pose is generated by summing two components: a coarse image directly predicted by a rendering network, and a warped texture image. While the warping itself is also predicted by the rendering network, the texture is estimated at the time of avatar creation and is static at runtime.
To enable the few-shot capability, we use the meta-learning stage on a dataset of videos, where we (meta)-train the inference (rendering) network, the embedding network, as well as the texture generation network. 

The separation of the target frames into two layers allows us both to improve the effective resolution and the speed of neural rendering. This is because we can use off-line avatar generation stage to synthesize high-resolution texture, while at test time both the first component (coarse image) and the warping of the texture need not contain high frequency details and can therefore be predicted by a relatively small rendering network. These advantages of our system are validated by extensive comparisons with previously proposed neural avatar systems. We also report on the smartphone-based real-time implementation of our system, which was beyond the reach of previously proposed models.

\section{Related work}

As discussed above, methods for the neural synthesis of realistic talking head sequences can be divided into many-shot (i.e.~requiring a video or multiple videos of the target person for learning the model)~\cite{Isola17,Kim18,Wang18c,Lombardi18} and a more recent group of few-shot/singe-shot methods capable of acquiring the model of a person from a single or a handful photographs~\cite{Wiles18,Zakharov19,Tripathy19,Ha19,Siarohin19,Wang19}. Our method falls into the latter category as we focus on the one-shot scenario (modeling from a single photograph).

Along another dimension, these methods can be divided according to the architecture of the generator network. Thus, several methods~\cite{Kim18,Wang18c, Zakharov19,Tripathy19} use generators based on \textit{direct synthesis}, where the image is generated using a sequence of convolutional operators, interleaved with elementwise non-linearities, and normalizations. Person identity information may be injected into such architecture, either with a lengthy learning process (in the many-shot scenario)~\cite{Kim18,Wang18c} or by using adaptive normalizations conditioned on person embeddings~\cite{Zakharov19,Tripathy19,Fu19}. The method~\cite{Zakharov19} effectively combines both approaches by injecting identity through adaptive normalizations, and then fine-tuning the resulting generator on the few-shot learning set. The direct synthesis approach for human heads can be traced back to \cite{Suwajanakorn17} that generated lips of a famous person in the talking head sequence, and further towards first works on conditional convolutional neural synthesis of generic objects such as~\cite{Dosovitskiy15}.

The alternative to the direct image synthesis is to use differentiable warping~\cite{Jaderberg15} inside the architecture. The X2Face approach~\cite{Wiles18} applies warping twice, first from the source image to a standardized image (texture), and then to the target image. The Codec Avatar system \cite{Lombardi18} synthesizes a pose-dependent texture for a simplified mesh geometry. The MarioNETte system~\cite{Ha19} applies warping to the intermediate feature representations. The Few-shot Vid-to-Vid system~\cite{Wang19} combines direct synthesis with the warping of the previous frame in order to obtain temporal continuity. The First Order Motion Model~\cite{Siarohin19} learns to warp the intermediate feature representation of the generator based on keypoints that are learned from data. Beyond heads, differentiable warping/texturing have recently been used for full body re-rendering~\cite{Neverova18,Shysheya19}. Earlier, DeepWarp system~\cite{Ganin16} used neural warping to alter the appearance of eyes for the purpose of gaze redirection, and \cite{Zhou16} also used neural warping for the resynthesis of generic scenes. 
Our method combines direct image synthesis with warping in a new way, as we obtain the fine layer by warping an RGB \textit{pose-independent} texture, while the coarse-grained \textit{pose-dependent} RGB component is synthesized by a neural network directly. 








\def\x{\mathbf{x}}
\def\y{\mathbf{y}}
\def\w{\mathbf{\omega}}
\def\m{\mathbf{m}}
\def\s{\mathbf{s}}
\def\e{\hat{\mathbf{e}}}
\def\f{\mathbf{f}}

\section{Methods}

We use video sequences annotated with keypoints and, optionally, segmentation masks, for training. We denote $t$-th frame of the $i$-th video sequence as $\x^i(t)$, corresponding keypoints as $\y^i(t)$, and segmentation masks as $\m^i(t)$
We will use an index $t$ to denote a target frame, and $s$ -- a source frame. Also, we mark all tensors, related to generated images, with a hat symbol, ex. $\hat\x^i(t)$. We assume the spatial size of all frames to be constant and denote it as $H \times W$. In some modules, input keypoints are encoded as an RGB image, which is a standard approach in a large body of previous works~\cite{Ha19, Wang19, Zakharov19}. In this work, we will call it a \textit{landmark} image. But, contrary to these approaches, at test-time we input the keypoints into the inference generator directly as a vector. This allows us to significantly reduce the inference time of the method.

\begin{figure}[t]
    \centering
    \includegraphics[width=\textwidth]{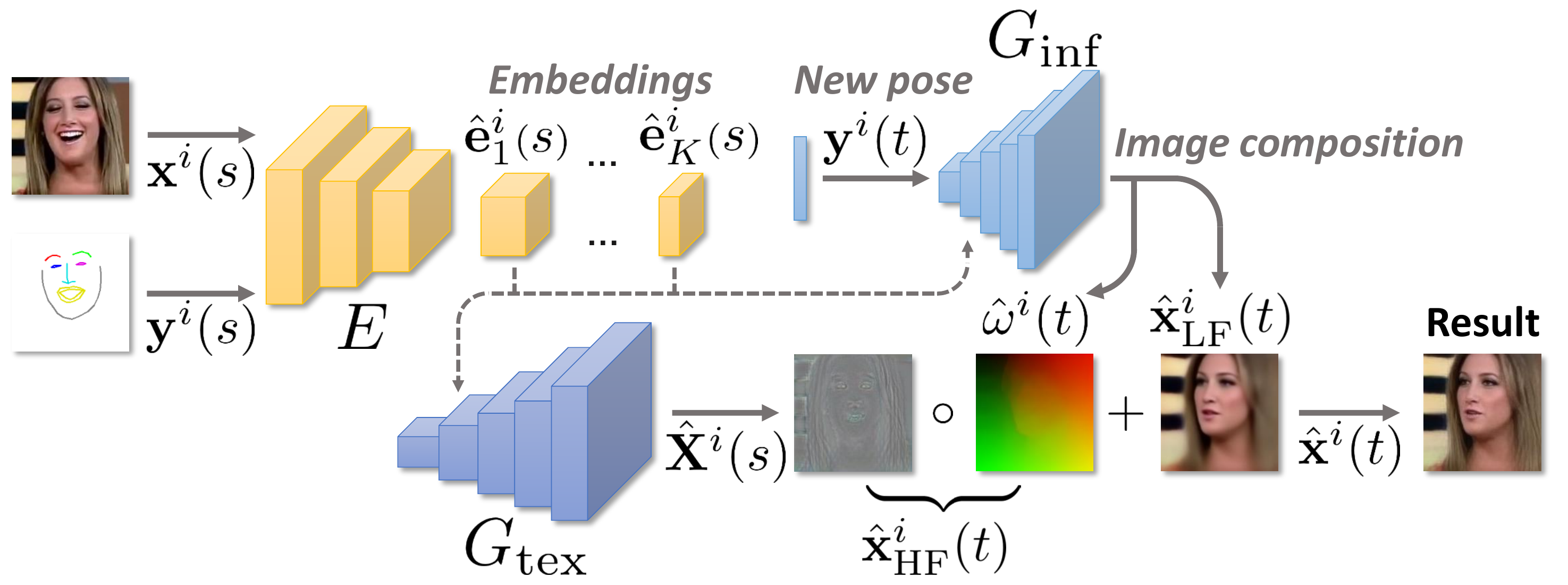}
    \caption{During training, we first encode a source frame into the embeddings, then we initialize adaptive parameters of both inference and texture generators, and predict a high-frequency texture. These operations are only done once per avatar. Target keypoints are then used to predict a low-frequency component of the output image and a warping field, which, applied to the texture, provides the high-frequency component. Two components are then added together to produce an output.}\label{fig:method_scheme}
\end{figure}

\subsection{Architecture}

In our approach, the following networks are trained in an end-to-end fashion:

\begin{itemize}
    \item The \textit{embedder} network $E \big( \x^i(s), \y^i(s) \big)$ encodes a concatenation of a source image and a landmark image into a stack of embeddings $\{ \e_k^i {\scriptstyle  (} s {\scriptstyle  )} \}$, which are used for initialization of the adaptive parameters inside the generators.
    
    \item The \textit{texture generator} network $G_\text{tex} \big( \{ \e_k^i {\scriptstyle  (} s {\scriptstyle  )} \} \big)$ initializes its adaptive parameters from the embeddings and decodes an inpainted high-frequency component of the source image, which we call a texture $\hat{\mathbf{X}}^i(s)$.
    
    \item The \textit{inference generator} network $G \big( \y^i(t), \{ \e_k^i {\scriptstyle  (} s {\scriptstyle  )} \} \big)$ maps target poses into a predicted image $\hat\x^i(t)$. The network accepts vector keypoints as an input and outputs a low-frequency layer of the output image $\hat\x_\text{LF}^i(t)$, which encodes basic facial features, skin color and lighting, and $\hat\w^i(t)$ -- a mapping between coordinate spaces of the texture and the output image. Then, the high-frequency layer of the output image is obtained by warping the predicted texture: $\hat\x_\text{HF}^i(t) = \hat\w^i(t) \circ \hat{\mathbf{X}}^i(s)$, and is added to a low-frequency component to produce the final image:
    
    
    \begin{equation}\label{eq:image}
        \hat\x^i(t) = \hat\x_\text{LF}^i(t) + \hat\x_\text{HF}^i(t) \, .
    \end{equation}
    
    \item Finally, the \textit{discriminator} network $D \big( \x^i(t), \y^i(t) \big)$, which is a conditional~\cite{Mirza14} relativistic~\cite{Martineau19} PatchGAN~\cite{Isola17}, maps a real or a synthesised target image, concatenated with the target landmark image, into realism scores $\s^i(t)$.
\end{itemize}

During training, we first input a source image $\x^i(s)$ and a source pose $\y^i(s)$, encoded as a landmark image, into the embedder. The outputs of the embedder are $K$ tensors $\e^i_k (s) $, which are used to predict the adaptive parameters of the texture generator and the inference generator. A high-frequency texture $\hat{\mathbf{X}}^i(s)$ of the source image is then synthesized by the texture generator. Next, we input corresponding target keypoints $\y^i(t)$ into the inference generator, which predicts a low-frequency component of the output image $\hat{\x}^i_\text{LF}(t)$ directly and a high-frequency component $\hat{\x}^i_\text{HF}(t)$ by warping the texture with a predicted field $\hat{\omega}^i(t)$. Finally, the output image $\hat\x^i(t)$ is obtained as a sum of these two components.

It is important to note that while the texture generator is manually forced to generate only a high-frequency component of the image via the design of the loss functions, which is described in the next section, we do not specifically constrain it to perform texture inpainting for occluded head parts. This behavior is emergent from the fact that we use two different images with different poses for initialization and loss calculation.

\subsection{Training process}

We use multiple loss functions for training. The main loss function responsible for the realism of the outputs is trained in an adversarial way~\cite{Goodfellow14}.  We also use pixelwise loss to preserve source lightning conditions and perceptual~\cite{Johnson16} loss to match the source identity in the outputs. Finally, a regularization of the texture mapping adds robustness to the random initialization of the model.

\subsubsection{Pixelwise and perceptual losses}

 ensure that the predicted images match the ground truth, and are respectively applied to low- and high-frequency components of the output images. Since usage of pixelwise losses assumes independence of all pixels in the image, the optimization process leads to blurry images \cite{Isola17}, which is suitable for the low-frequency component of the output. Thus the pixelwise loss is calculated by simply measuring mean $L_1$ distance between the target image and the low-frequency component:

\begin{equation}\label{eq:loss_pix}
    \mathcal{L}^G_\text{pix} = \frac{1}{HW} || \hat\x^i_\text{LF}(t) - \x^i(t) ||_1 \, .
\end{equation}

On the contrary, the optimization of the perceptual loss leads to crisper and more realistic images \cite{Johnson16}, which we utilize to train the high-frequency component. 
To calculate the perceptual loss, we use the stop-gradient operator $\text{SG}$, which allows us to prevent the gradient flow into a low-frequency component. The input generated image is, therefore, calculated as following:

\begin{equation}\label{eq:loss_perceptual_sg}
    \tilde\x^i(t) = \text{SG} \big( \hat\x^i_\text{LF}(t) \big) + \hat\x^i_\text{HF}(t) \, .
\end{equation}

Following \cite{Ha19} and \cite{Zakharov19}, our variant of the perceptual loss consists of two components: features evaluated using an ILSVRC (ImageNet) pre-trained VGG19 network \cite{Simonyan14}, and the VGGFace network \cite{Parkhi15}, trained for face recognition. If we denote the intermediate features of these networks as $\f^i_{k,\text{IN}}(t)$ and $\f^i_{k,\text{face}}(t)$, and their spatial size as $H_k \times W_k$, the objectives can be written as follows:

\begin{equation}\label{eq:loss_in}
    \mathcal{L}_\text{IN}^G = \frac{1}{K} \sum_k \frac{1}{H_k W_k} || \tilde\f^i_{k,\text{IN}}(t) - \f^i_{k,\text{IN}}(t) ||_1 \, ,
\end{equation}

\begin{equation}\label{eq:loss_face}
    \mathcal{L}_\text{face}^G = \frac{1}{K} \sum_k \frac{1}{H_k W_k} || \tilde\f^i_{k,\text{face}}(t) - \f^i_{k,\text{face}}(t) ||_1 \, .
\end{equation}

\subsubsection{Texture mapping regularization}

is proposed to improve the stability of the training. In our model, the coordinate space of the texture is learned implicitly, and there are two degrees of freedom that can mutually compensate each other: the position of the face in the texture, and the predicted warping. If, after initial iterations, the major part of the texture is left unused by the model, it can easily compensate that with a more distorted warping field. This artifact of an initialization is not fixed during training, and clearly is not the behavior we need, since we want all the texture to be used to achieve the maximum effective resolution in the outputs. We address the problem by regularizing the warping in the first iterations to be close to an identity mapping:

\begin{equation}\label{eq:loss_reg}
    \mathcal{L}^G_\text{reg} = \frac{1}{HW} ||\w^i(t) - \mathcal{I}||_1 \, .
\end{equation}

\subsubsection{Adversarial loss}

 is optimized by both generators, the embedder and the discriminator networks. Usually, it resembles a binary classification loss function between real and fake images, which discriminator is optimized to minimize, and generators -- maximize~\cite{Goodfellow14}. We follow a large body of previous works~\cite{Brock19, Wang19, Ha19, Zakharov19} and use a hinge loss as a substitute for the original binary cross entropy loss. We also perform relativistic realism score calculation~\cite{Martineau19}, following its recent success in tasks such as super-resolution~\cite{Wang18c} and denoising~\cite{Kim19}. Additionally, we use PatchGAN~\cite{Isola17} formulation of the adversarial learning. The discriminator is trained only with respect to its adversarial loss $\mathcal{L}_\text{adv}^D$, while the generators and the embedder are trained via the adversarial loss $\mathcal{L}_\text{adv}^G$, and also a feature matching loss $\mathcal{L}_\text{FM}$~\cite{Wang18b}. The latter is introduced for better stability of the training.

\subsection{Texture enhancement}

\begin{figure}[t]
    \centering
    \includegraphics[width=\textwidth]{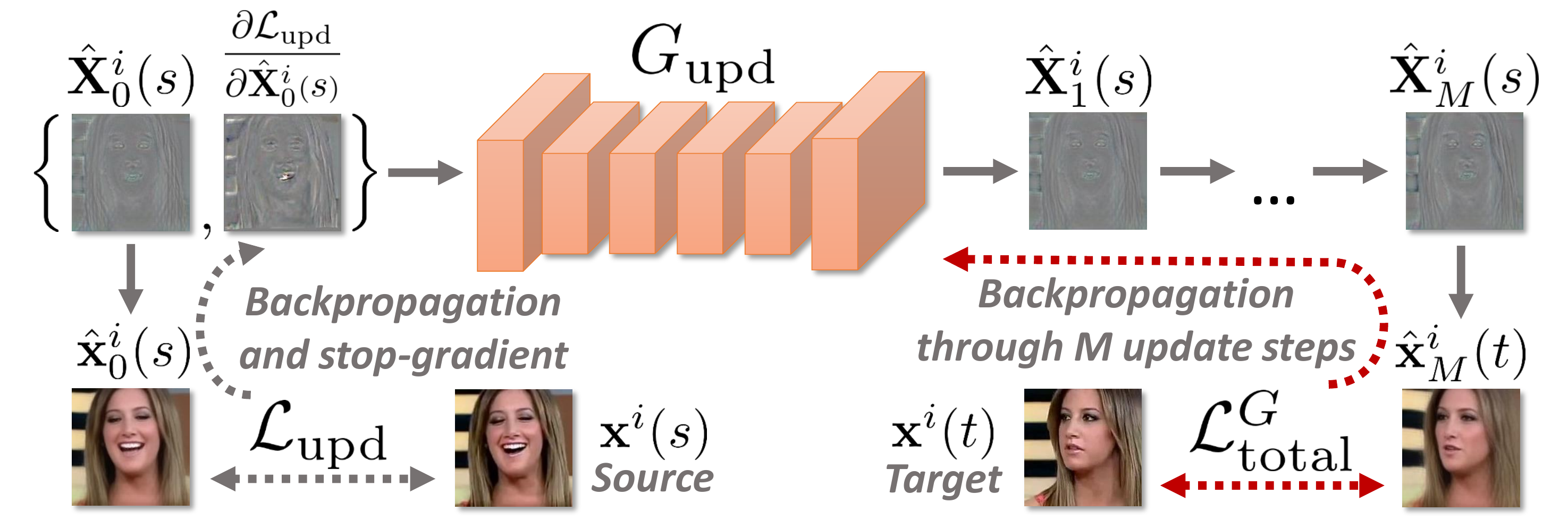}
    \caption{Texture enhancement network (updater) accepts the current state of the texture and the guiding gradients to produce the next state. The guiding gradients are obtained by reconstructing the source image from the current state of the texture and matching it to the ground-truth via a lightweight updater loss. These gradients are only used as inputs and are detached from the computational graph. This process is repeated M times. The final state of the texture is then used to obtain a target image, which is matched to the ground-truth via the same loss as the one used during training of the main model. The gradients from this loss are then backpropagated through all M copies of the updater network.}\label{fig:updater_scheme}
\end{figure}

To minimize the identity gap, \cite{Zakharov19} suggested to fine-tune the generator weights to the few-shot training set. Training on a person-specific source data leads to significant improvement in realism and identity preservation of the synthesized images~\cite{Zakharov19}, but is computationally expensive. Moreover, when the source data is scarce, like in one-shot scenario, fine-tuning may lead to over-fitting and performance degradation, which is observed in~\cite{Zakharov19}. We address both of these problems by using a learned gradient descend~(LGD) method~\cite{Andrychowicz16} to optimize only the synthesized texture $\hat{\mathbf{X}}^i(s)$. Optimizing with respect to the texture tensor prevents the model from overfitting, while the LGD allows us to perform optimization with respect to computationally expensive objectives by doing forward passes through a pre-trained network.

Specifically, we introduce a lightweight loss function $\mathcal{L}_\text{upd}$ (we use a sum of squared errors), that measures the distance between a generated image and a ground-truth in the pixel space, and a \textit{texture updating} network $G_\text{upd}$, that uses the current state of the texture and the gradient of $\mathcal{L}_\text{upd}$ with respect to the texture to produce an update $\Delta \hat{\mathbf{X}}^i(s)$. During fine-tuning we perform $M$ update steps, each time measuring the gradients of $\mathcal{L}_\text{upd}$ with respect to an updated texture. The visualization of the process can be seen in Figure~\ref{fig:updater_scheme}. More formally, each update is computed as:

\begin{equation}
    \hat{\mathbf{X}}^i_{m+1}(s) = \hat{\mathbf{X}}^i_m(s) +  G_\text{upd} \Bigg( \hat{\mathbf{X}}^i_m(s), \dfrac{ \partial \mathcal{L}_\text{upd} }{ \partial \hat{\mathbf{X}}^i_m {\scriptstyle (} s {\scriptstyle )} } \Bigg) \, ,
\end{equation}

where $m \in \{0, \dots, M-1 \}$ denotes an iteration number, with $\hat{\mathbf{X}}^i_0(s) \equiv \hat{\mathbf{X}}^i(s)$. 

The network $G_\text{upd}$ is trained by back-propagation through all $M$ steps. For training, we use the same objective $\mathcal{L}^G_\text{total}$ that was used during the training of the base model. We evaluate it using a target frame $\x^i(t)$ and a generated frame

\begin{equation} \label{eq:update}
    \hat\x^i_M(t) = \hat\x^i_\text{LF}(t) + \hat\omega^i(t) \circ \hat{\mathbf{X}}^i_M{(s)} \, .
\end{equation}

It is important to highlight that $\mathcal{L}_\text{upd}$ is not used for training of $G_\text{upd}$, but simply guides the updates to the texture. Also, the gradients with respect to this loss are evaluated using the source image, while the objective in Eq.~\ref{eq:update} is calculated using the target image, which implies that the network has to produce updates for the whole texture, not just a region ``visible'' on the source image. Lastly, while we do not propagate any gradients into the generator part of the base model, we keep training the discriminator using the same objective $\mathcal{L}^D_\text{adv}$. Even though training the updater network jointly with the base generator is possible, and can lead to better quality (following the success of model agnostic meta-learning~\cite{Finn17} method), we resort to two-stage training due to memory constraints.

\subsection{Segmentation}

The presence of static background leads to a certain degradation of our model for two reasons. Firstly, part of the capacity of the texture and the inference generators has to be spent on modeling high variety of background patterns. Secondly, and more importantly, the static nature of backgrounds in most training videos biases the warping towards identity mapping. We therefore, have found it advantageous to include background segmentation into our model.

We use a state-of-the-art face and body segmentation model~\cite{Gong19} to obtain the ground truth masks. Then, we add the mask prediction output $\hat\m^i(t)$ to our inference generator alongside with its other outputs, and train it via a binary cross-entropy loss $\mathcal{L}_\text{seg}$ to match the ground truth mask $\m^i(t)$. To filter out the training signal, related to the background, we have explored multiple options. Simple masking of the gradients that are fed into the generator leads to severe overfitting of the discriminator. We also could not simply apply the ground truth masks to all the images in the dataset, since the model~\cite{Gong19} works so well that it produces a sharp border between the foreground and the background, leading to border artifacts that emerge after adversarial training.

Instead, we have found out that masking the ground truth images that are fed to the discriminator with the predicted masks $\hat\m^i(t)$ works well. Indeed, these masks are smooth and prevent the discriminator from overfitting to the lack of background, or sharpness of the border. We do not backpropagate the signal from the discriminator and from perceptual losses to the generator via the mask pathway (i.e.\ we use stop gradient/detach operator $\text{SG}\big( \hat\m^i(t) \big)$ before applying the mask). The stop-gradient operator also ensures that the training does not converge to a degenerate state (empty foreground).

\subsection{Implementation details}

All our networks consist of pre-activation residual blocks~\cite{He16} with LeakyReLU activations. We set a minimum number of features in these blocks to 64, and a maximum to 512. By default, we use half the number of features in the inference generator, but we also evaluate our model with full- and quater-capacity inference part, with the results provided in the experiments section.

We use batch normalization~\cite{Ioffe15} in all the networks except for the embedder and the texture updater. Inside the texture generator, we pair batch normalization with adaptive SPADE layers~\cite{Wang19}. We modify these layers to predict pixelwise scale and bias coefficients using feature maps, which are treated as model parameters, instead of being input from a different network. This allows us to save memory by removing additional networks and intermediate feature maps from the optimization process, and increase the batch size. Also, following~\cite{Wang19}, we predict weights for all $1 \times 1$ convolutions in the network from the embeddings $\{ \e_k^i {\scriptstyle  (} s {\scriptstyle  )} \}$, which includes the scale and bias mappings in AdaSPADE layers, and skip connections in the residual upsampling blocks. In the inference generator, we use standard adaptive batch normalization layers~\cite{Brock19}, but also predict weights for the skip connections from the embeddings. 

We do simultaneous gradient descend on parameters of the generator networks and the discriminator using Adam~\cite{Diederik14} with a learning rate of $2\cdot10^{-4}$. We use 0.5 weight for adversarial losses, and 10 for all other losses, except for the VGGFace perceptual loss (Eq. \ref{eq:loss_face}), which is set to 0.01. The weight of the regularizer (Eq. \ref{eq:loss_reg}) is then multiplicatively reduced by 0.9 every 50 iterations. We train our models on 8 NVIDIA P40 GPUs with the batch size of 48 for the base model, and a batch size of 32 for the updater model. We set unrolling depth $M$ of the updater to 4 and use a sum of squared errors as the lightweight objective. Batch normalization statistics are synchronized across all GPUs during training. During inference they are replaced with ``standing'' statistics, similar to~\cite{Brock19}, which significantly improves the quality of the outputs, compared to the usage of running statistics. Spectral normalization is also applied in all linear and convolutional layers of all networks.

Please refer to the supplementary material for a detailed description of our model's architecture, as well as the discussion of training and architectural features that we have adopted.

\section{Experiments}

We perform evaluation in multiple scenarios. First, we use the original VoxCeleb2~\cite{Chung18} dataset to compare with state-of-the-art systems. To do that, we annotated this dataset using an off-the-shelf facial landmarks detector~\cite{Bulat17}. Overall, the dataset contains 140697 videos of 5994 different people. We also use a high-quality version of the same dataset, additionally annotated with the segmentation masks (which were obtained using a model~\cite{Gong19}), to measure how the performance of our model scales with a dataset of a significantly higher quality. We obtained this version by downloading the original videos via the links provided in the VoxCeleb2 dataset, and filtering out the ones with low resolution. This dataset is, therefore, significantly smaller and contains only 14859 videos of 4242 people, with each video having at most 250 frames (first 10 seconds). Lastly, we do ablation studies on both VoxCeleb2 and VoxCeleb2-HQ, and report on a smartphone-based implementation of the method. For comparisons and ablation studies we show the results qualitatively and also evaluate the following metrics:

\begin{itemize}
    \item Learned perceptual image patch similarity~\cite{Zhang18b} (LPIPS), which measures overall predicted image similarity to ground truth.
    \item Cosine similarity between the embedding vectors of a state-of-the-art face recognition network~\cite{Deng19} (CSIM), calculated using the synthesized and the target images. This metric evaluates the identity mismatch.
    \item Normalized mean error of the head pose in the synthesized image (NME). We use the same network~\cite{Bulat17}, which was used for the annotation of the dataset, to evaluate the pose of the synthesized image. We normalize the error, which is a mean euclidean distance between the predicted and the target points, by the distance between the eyes in the target pose, multiplied by 10.
    \item Multiply-accumulate operations (MACs), which measure the complexity of each method. We exclude from the evaluation initialization steps, which are calculated only once per avatar.
\end{itemize}

The test set in both datasets does not intersect with the train set in terms of videos or identities. For evaluation, we use a subset of 50 test videos with different identities (for VoxCeleb2, it is the same as in~\cite{Zakharov19}). The first frame in each sequence is used as a source. Target frames are taken sequentially at 1 FPS.

We only discuss most important results in the main paper. For additional qualitative results and comparisons please refer to the supplementary materials.

\subsection{Comparison with the state-of-the-art methods}

\begin{figure}[t]
    \centering
    \includegraphics[width=\textwidth]{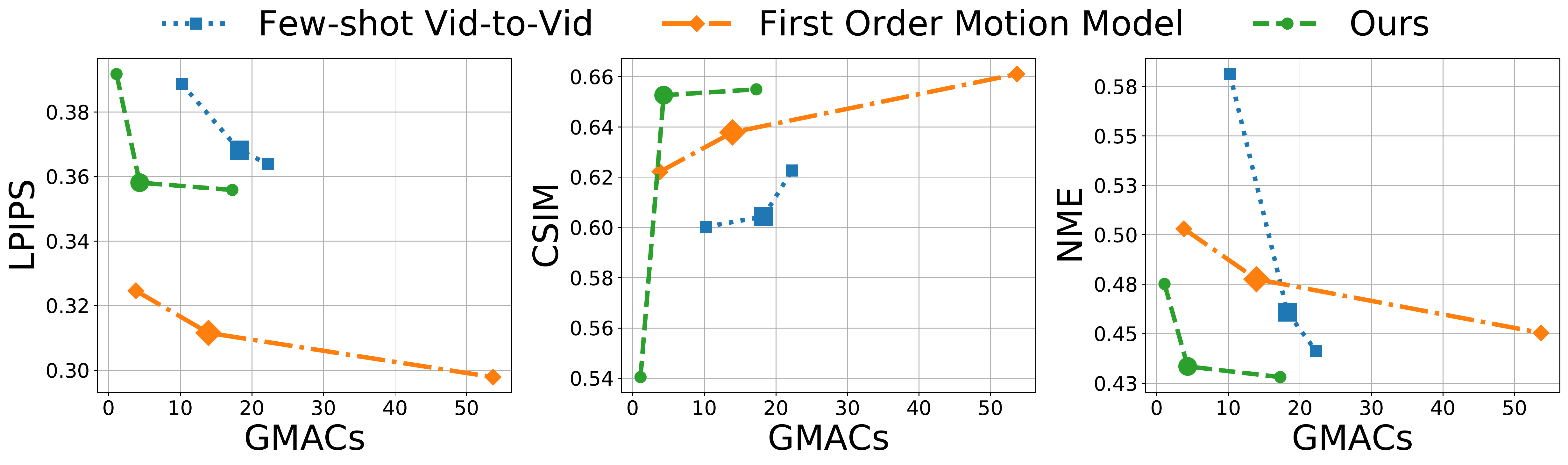}
    \caption{In order to evaluate a quality against performance trade off, we train a family of models with varying complexity for each of the compared methods. For quality metrics, we have compared synthesized images to their targets using a perceptual image similarity (LPIPS\,$\downarrow$), identity preservation metric (CSIM\,$\uparrow$), and a normalized pose error (NME\,$\downarrow$). We highlight a model which was used for the comparison in Figure~\ref{fig:vc2_comp} with a bold marker. We observe that our model outperforms the competitors in terms of identity preservation (CSIM) and pose matching (NME) in the settings, when models' complexities are comparable. In order to better compare with FOMM, we did a user study, where users have preferred the image generated by our model to FOMM 59.6\% of the time.}\label{fig:vc2_quant_comp}
\end{figure}

We compare against three state-of-the-art systems: Few-shot Talking Heads~\cite{Zakharov19}, Few-shot Vid-to-Vid~\cite{Wang19} and First Order Motion Model~\cite{Siarohin19}. The first system is a problem-specific model designed for avatar creation. Few-shot Vid-to-Vid is a state-of-the-art video-to-video translation system, which has also been successfully applied to this problem. First Order Motion Model (FOMM) is a general motion transfer system that does not use precomputed keypoints, but can also be used as an avatar. We believe that these models are representative of the most recent and successful approaches to one-shot avatar generation. We also acknowledge the work of~\cite{Ha19}, but do not compare to them extensively due to unavailability of the source code, pretrained models or pre-calculated results. A small-scale qualitative comparison is provided in the supplementary materials. Additionally, their method is limited to the usage of 3D keypoints, while our method does not have such restriction. Lastly, since Few-shot Vid-to-Vid is an autoregressive model, we use a full test video sequence for evaluation (25 FPS) and save the predicted frames at 1 FPS.

\begin{figure}[t]
    \centering    
    \setlength{\wid}{0.155\textwidth}
    \begin{tabular}{cccccc}
        \includegraphics[align=c,bmargin=0.1cm,width=\wid]{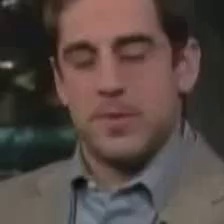} &
        \includegraphics[align=c,width=\wid]{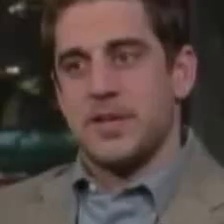} &
        \includegraphics[align=c,width=\wid]{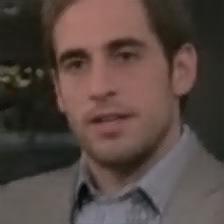} &
        \includegraphics[align=c,width=\wid]{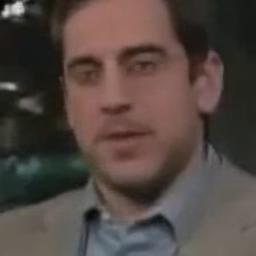} &
        \includegraphics[align=c,width=\wid]{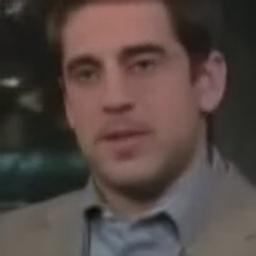} &
        \includegraphics[align=c,width=\wid]{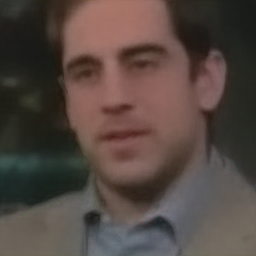} \\ 
        \includegraphics[align=c,bmargin=0.1cm,width=\wid]{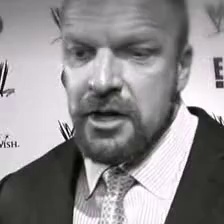} &
        \includegraphics[align=c,width=\wid]{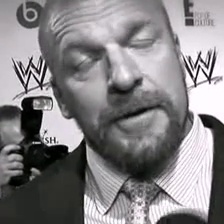} &
        \includegraphics[align=c,width=\wid]{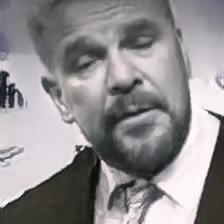} &
        \includegraphics[align=c,width=\wid]{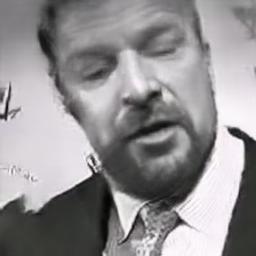} &
        \includegraphics[align=c,width=\wid]{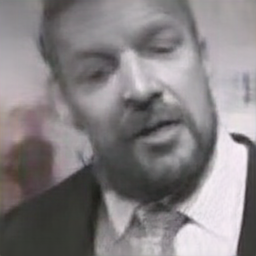} &
        \includegraphics[align=c,width=\wid]{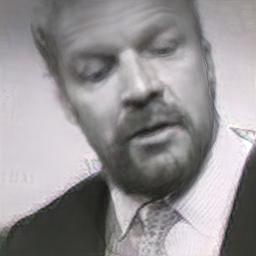} \\ 
        \includegraphics[align=c,bmargin=0.1cm,width=\wid]{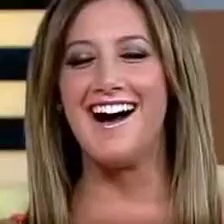} &
        \includegraphics[align=c,width=\wid]{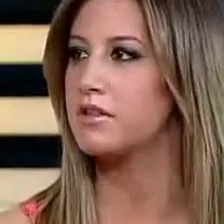} &
        \includegraphics[align=c,width=\wid]{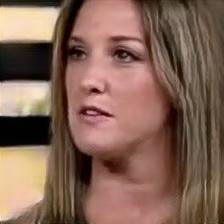} &
        \includegraphics[align=c,width=\wid]{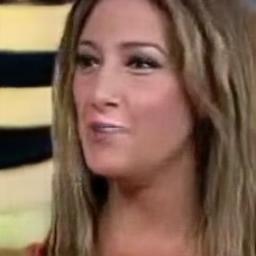} &
        \includegraphics[align=c,width=\wid]{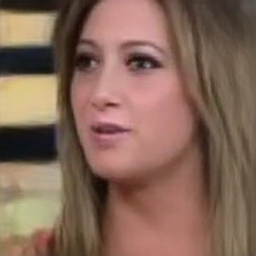} &
        \includegraphics[align=c,width=\wid]{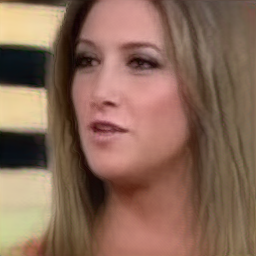} \\ 
        \includegraphics[align=c,bmargin=0.1cm,width=\wid]{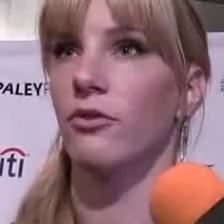} &
        \includegraphics[align=c,width=\wid]{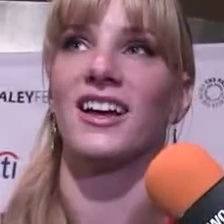} &
        \includegraphics[align=c,width=\wid]{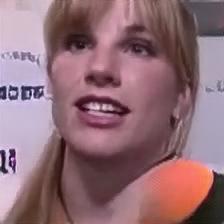} &
        \includegraphics[align=c,width=\wid]{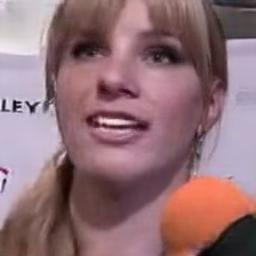} &
        \includegraphics[align=c,width=\wid]{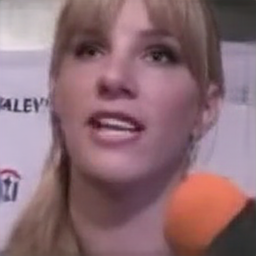} &
        \includegraphics[align=c,width=\wid]{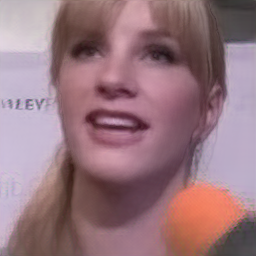} \\ 
        \textbf{Source} & \textbf{Target} & \textbf{\begin{tabular}{c} Few-shot \\ T. Heads \end{tabular}} & \textbf{\begin{tabular}{c} Few-shot \\ Vid-to-Vid \end{tabular}} & \textbf{FOMM} & \textbf{Ours}
    \end{tabular}
    \caption{Comparison on a VoxCeleb2 dataset. The task is to reenact a \textbf{target} image, given a \textbf{source} image and target keypoints. The compared methods are Few-shot Talking Heads~\cite{Zakharov19}, Few-shot Vid-to-Vid~\cite{Wang19}, First Order Motion Model (FOMM)~\cite{Siarohin19} and our proposed Bi-layer Model. For each method, we used the models with a similar number of parameters, and picked source and target images to have diverse poses and expressions, in order to highlight the differences between the compared methods.}
    \label{fig:vc2_comp}
\end{figure}

Importantly, the base models in these approaches have a lot of computational complexity, so for each method we evaluate a family of models by varying the number of parameters. The performance comparison for each family is reported in Figure~\ref{fig:vc2_quant_comp} (with Few-shot Talking Heads being excluded from this evaluation, since their performance is much worse than the compared methods). Overall, we can see that our model's family outperforms competing methods in terms of pose error and identity preservation, while being, on average, up to an order of magnitude faster. To better compare with FOMM in terms of image similarity, we have performed a user study, where we asked crowd-sourced users which generated image better matches the ground truth. In total, 361 users evaluated 1600 test pairs of images, with each one seeing on average 21 pairs. In 59.6\% of comparisons, the result of our medium model was preferred to a medium sized model of FOMM.

Another important note is on how the complexity was evaluated. In Few-shot Vid-to-Vid we have additionally excluded from the evaluation parts that are responsible for the temporal consistency, since other compared methods are evaluated frame-by-frame and do not have such overhead. Also, in FOMM we have excluded the keypoints extractor network, because this overhead is shared implicitly by all the methods via usage of the precomputed keypoints.

We visualize the results for medium-sized models of each of the compared methods in Figure~\ref{fig:vc2_comp}. Since all methods perform similarly in case when source and target images have marginal differences, we have shown the results where a source and a target have different head poses. In this extrapolation setting, our method has a clear advantage, while other methods either introduce more artifacts or more blurriness.

\subsubsection{Evaluation on high-quality images.}

\begin{figure}[t]
    \centering    
    \setlength{\wid}{0.155\textwidth}
    \begin{tabular}{cccccc}
        \includegraphics[align=c,width=\wid,bmargin=0.1cm]{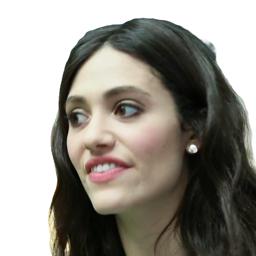} & 
        \includegraphics[align=c,width=\wid]{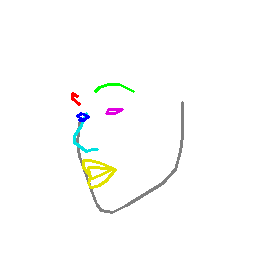} &
        \includegraphics[align=c,width=\wid]{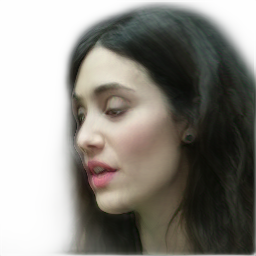} &
        \includegraphics[align=c,width=\wid]{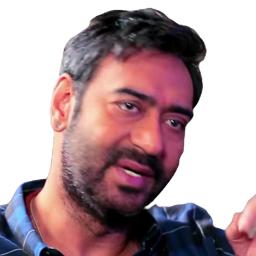} & 
        \includegraphics[align=c,width=\wid]{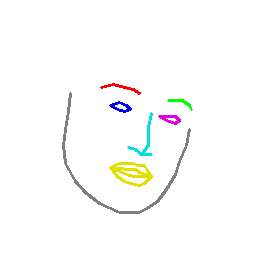} &
        \includegraphics[align=c,width=\wid]{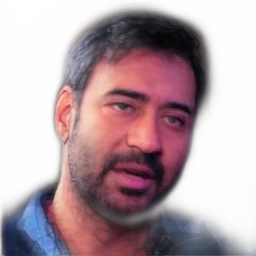} \\
        \includegraphics[align=c,width=\wid,bmargin=0.1cm]{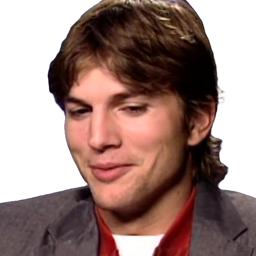} & 
        \includegraphics[align=c,width=\wid]{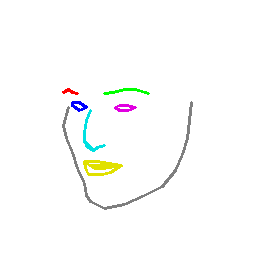} &
        \includegraphics[align=c,width=\wid]{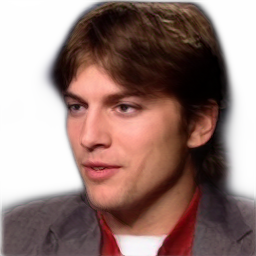} &
        \includegraphics[align=c,width=\wid]{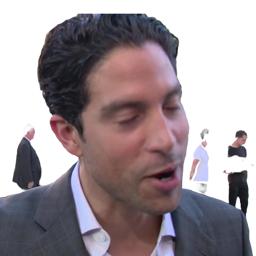} & 
        \includegraphics[align=c,width=\wid]{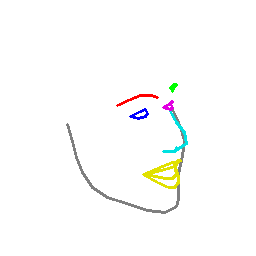} &
        \includegraphics[align=c,width=\wid]{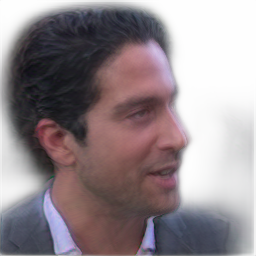} \\ 
        \includegraphics[align=c,width=\wid,bmargin=0.1cm]{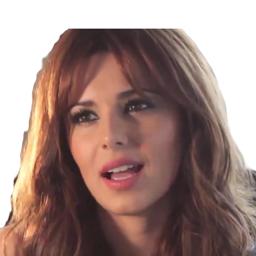} & 
        \includegraphics[align=c,width=\wid]{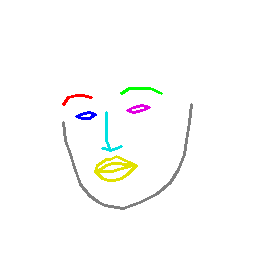} &
        \includegraphics[align=c,width=\wid]{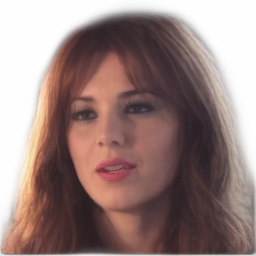} &
        \includegraphics[align=c,width=\wid]{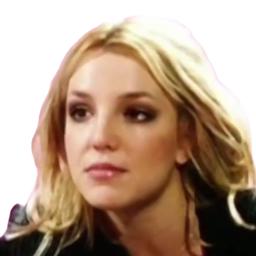} & 
        \includegraphics[align=c,width=\wid]{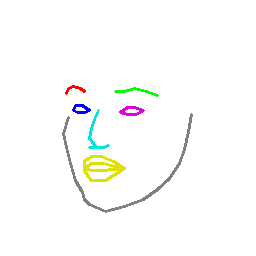} &
        \includegraphics[align=c,width=\wid]{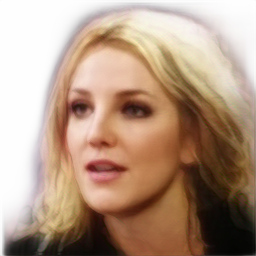} \\
        \includegraphics[align=c,width=\wid,bmargin=0.1cm]{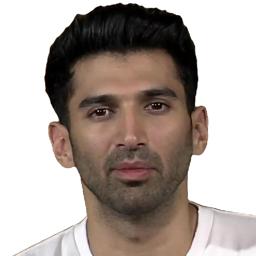} & 
        \includegraphics[align=c,width=\wid]{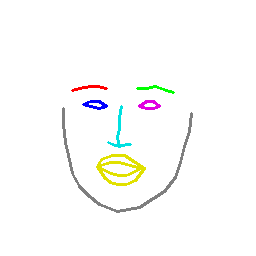} &
        \includegraphics[align=c,width=\wid]{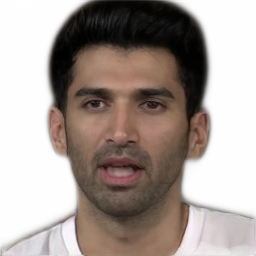} &
        \includegraphics[align=c,width=\wid]{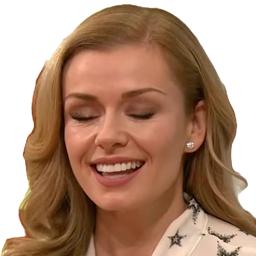} & 
        \includegraphics[align=c,width=\wid]{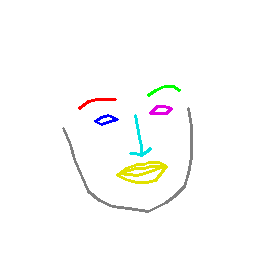} &
        \includegraphics[align=c,width=\wid]{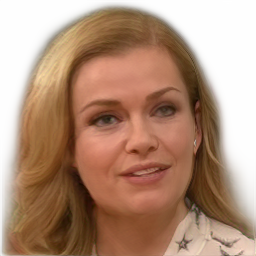} \\
        \textbf{Source} & \textbf{Pose} & \textbf{Result} & \textbf{Source} & \textbf{Pose} & \textbf{Result}
    \end{tabular}
    \caption{High quality synthesis results. We can see that our model is both capable of viewpoint extrapolation and low identity gap synthesis. The architecture in this experiment has the same number of parameters as the medium architecture in the previous comparison.}
    \label{fig:hq_results}
\end{figure}

Next, we evaluate our method on the high-quality dataset and present the results in Figure~\ref{fig:hq_results}. Overall, in this case, our method is able to achieve a smaller identity gap, compared to the dataset with the background. We also show the decomposition between the texture and a low frequency component in~Figure~\ref{fig:hq_detailed_results}. Lastly, in Figure~\ref{fig:reenactment}, we show that our texture enhancement pipeline allows us to render small person-specific features like wrinkles and moles on out-of-domain examples. For more qualitative examples, as well as reenactment examples with a driver of a different person, please refer to the supplementary materials.

\begin{table}[t]
    \parbox{.49\linewidth}{
        \centering
        \setlength{\wid}{0.11\textwidth}
        \begin{tabular}{cccc}
            \includegraphics[align=c,width=\wid,bmargin=0.1cm]{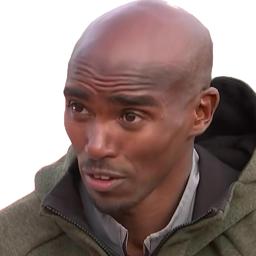} & 
            \includegraphics[align=c,width=\wid]{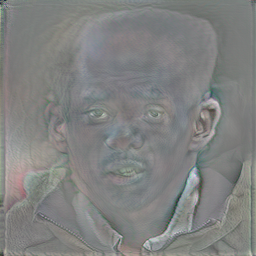} &
            \includegraphics[align=c,width=\wid]{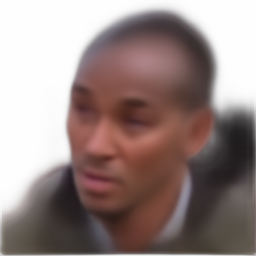} &
            \includegraphics[align=c,width=\wid]{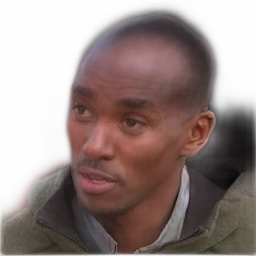} \\
            \includegraphics[align=c,width=\wid,bmargin=0.1cm]{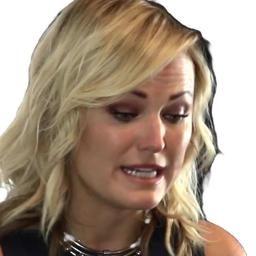} & 
            \includegraphics[align=c,width=\wid]{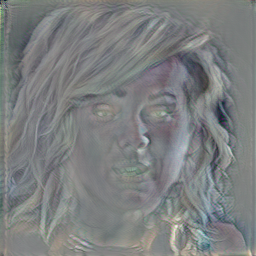} &
            \includegraphics[align=c,width=\wid]{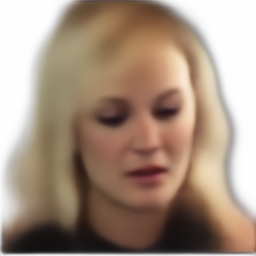} &
            \includegraphics[align=c,width=\wid]{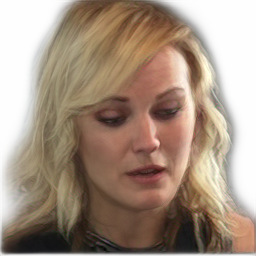} \\
            \textbf{Source} & \textbf{Texture} & \textbf{LF} & \textbf{Result}
        \end{tabular}
        \captionof{figure}{Detailed results on the generation process of the output image. \textbf{LF} denotes a low-frequency component.}\label{fig:hq_detailed_results}
    }
    \hfill
    \parbox{.49\linewidth}{
        \centering
        \setlength{\wid}{0.11\textwidth}
        \begin{tabular}{cccc}
            \includegraphics[align=c,width=\wid,bmargin=0.1cm]{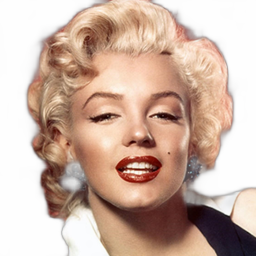} &
            \includegraphics[align=c,width=\wid]{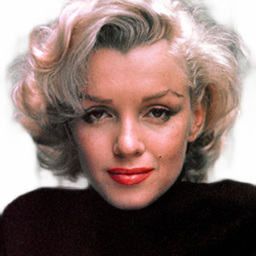} &
            \includegraphics[align=c,width=\wid]{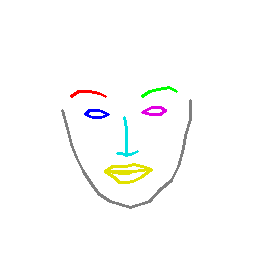} &
            \includegraphics[align=c,width=\wid]{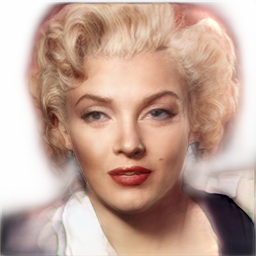} \\
            \includegraphics[align=c,width=\wid,bmargin=0.1cm]{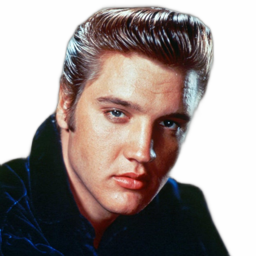} &
            \includegraphics[align=c,width=\wid]{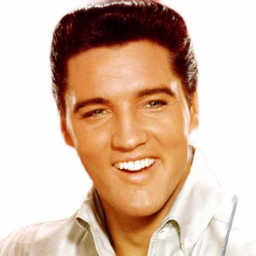} &
            \includegraphics[align=c,width=\wid]{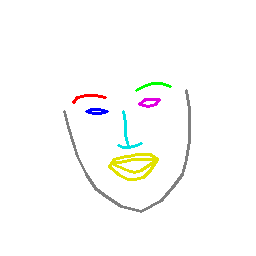} &
            \includegraphics[align=c,width=\wid]{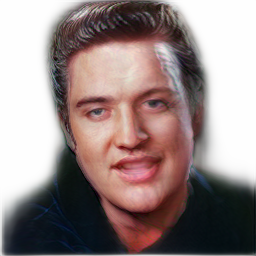} \\
            \textbf{Source} & \textbf{Target} & \textbf{Pose} & \textbf{Ours}
        \end{tabular}
        \captionof{figure}{Our method can preserve a lot of details in the facial features, like the famous Marylin's mole.}\label{fig:reenactment}
    }
\end{table}

\subsubsection{Smartphone-based implementation.}

We train our model using PyTorch~\cite{PyTorch} and then port it to smartphones with Qualcomm Snapdragon chips. There are several frameworks which provide APIs for mobile inference on such devices. From our experiments, we measured the Snapdragon Neural Processing Engine (SNPE)~\cite{SNPE} to be about 1.5 times faster than PyTorch Mobile~\cite{PyTorch} and up to two times faster than TensorFlow Lite~\cite{TFLite}. The medium-sized model ported to the Snapdragon 855 (Adreno 640 GPU, FP16 mode) takes 42 ms per frame, which is sufficient for real-time performance, given that the keypoint tracking is being run in parallel, e.g.\ on a mobile CPU.

\subsubsection{Ablation study.}

Finally, we evaluate the contribution of individual components. First, we evaluate the contribution of adaptive SPADE layers in the texture generator (by replacing them with adaptive batch normalization and per-pixel biases) and adaptive skip-connections in both generators. A model with these features removed makes up our baseline. Lastly, we evaluate the contribution of the updater network. The results can be seen in Table~\ref{table:ablation} and Figure~\ref{fig:ablation}. We evaluate the baseline approach only on a VoxCeleb2 dataset, while the full models with and without the updater network are evaluated on both low- and high-quality datasets. Overall, we see a significant contribution of each component with respect to all metrics, which is particularly noticeable in the high-quality scenario. In all ablation comparisons, medium-sized models were used.

\begin{table}[t]
    \parbox{.49\linewidth}{
        \centering
        \begin{tabular}{ l c c c c}
            Method \;&\; LPIPS\,$\downarrow$ \,&\, CSIM\,$\uparrow$ \,&\, NME\,$\downarrow$ \\
            \hline
            \multicolumn{4}{c}{VoxCeleb2} \\
            \hline
            Baseline & 0.377 & 0.547 & 0.447 \\
            Ours  & 0.370 & 0.595 & 0.441 \\
            +Updater & \textbf{0.358} & \textbf{0.653} & \textbf{0.433} \\
            \hline
            \multicolumn{4}{c}{VoxCeleb2-HQ} \\                                
            \hline
            Ours & 0.313 & 0.432 & 0.476 \\
            +Updater & \textbf{0.298} & \textbf{0.649} & \textbf{0.456}
        \end{tabular}
        \caption{Ablation studies of our approach. We first evaluate the baseline method without AdaSPADE or adaptive skip connections. Then we add these layers, following~\cite{Wang19}, and observe significant quality improvement. Finally, our updater network provides even more improvement across all metrics, especially noticeable in the high-quality scenario.}\label{table:ablation}
    }
    \hfill
    \parbox{.49\linewidth}{
        \centering
        \setlength{\wid}{0.11\textwidth}
        \begin{tabular}{cccc}
            \includegraphics[align=c,width=\wid,bmargin=0.1cm]{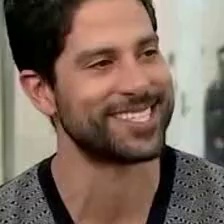} & 
            \includegraphics[align=c,width=\wid]{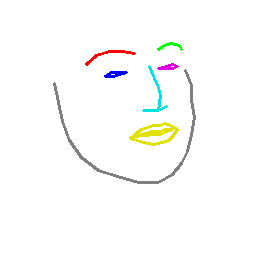} &
            \includegraphics[align=c,width=\wid]{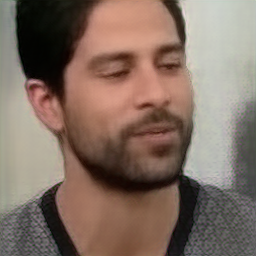} &
            \includegraphics[align=c,width=\wid]{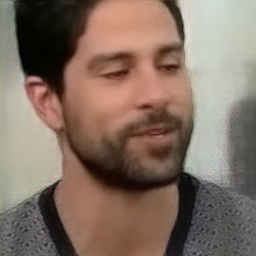} \\
            \includegraphics[align=c,width=\wid,bmargin=0.1cm]{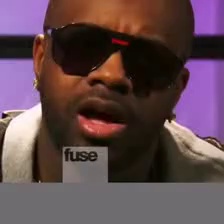} & 
            \includegraphics[align=c,width=\wid]{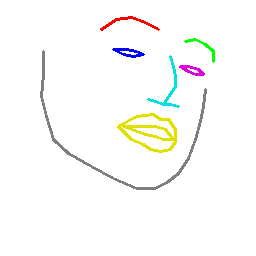} &
            \includegraphics[align=c,width=\wid]{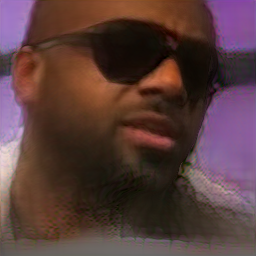} &
            \includegraphics[align=c,width=\wid]{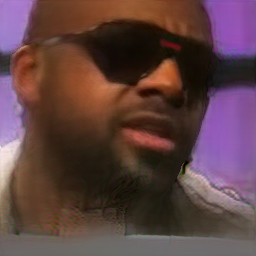} \\
            \includegraphics[align=c,width=\wid,bmargin=0.1cm]{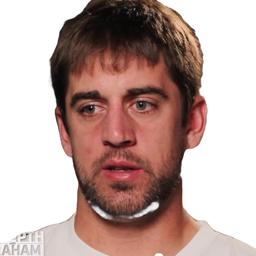} & 
            \includegraphics[align=c,width=\wid]{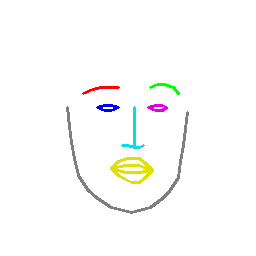} &
            \includegraphics[align=c,width=\wid]{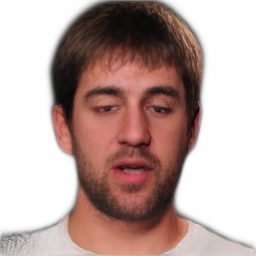} &
            \includegraphics[align=c,width=\wid]{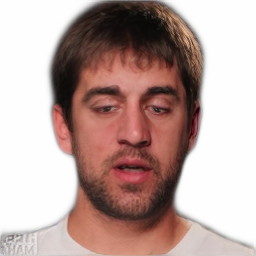} \\
            \includegraphics[align=c,width=\wid,bmargin=0.1cm]{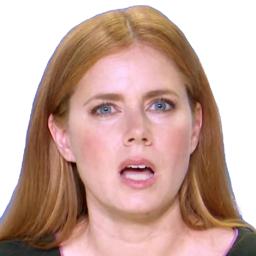} & 
            \includegraphics[align=c,width=\wid]{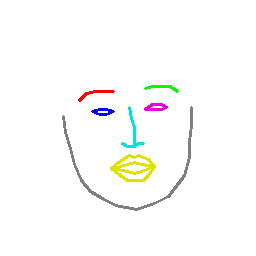} &
            \includegraphics[align=c,width=\wid]{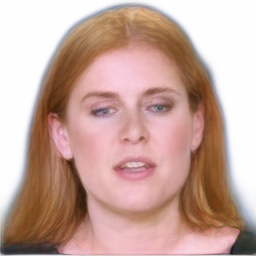} &
            \includegraphics[align=c,width=\wid]{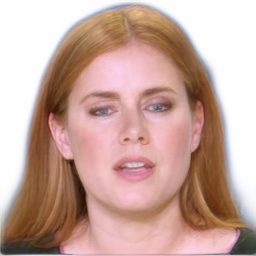} \\
            \textbf{Source} & \textbf{Pose} & \textbf{Ours} & \textbf{\begin{tabular}{c} +Upd. \end{tabular}}
        \end{tabular}
        \captionof{figure}{Examples from the ablation study on VoxCeleb2 (first two rows) and VoxCeleb2-HQ (last two rows).}\label{fig:ablation}
    }
\end{table}
\section{Conclusion}


We have proposed a new neural rendering-based system that creates head avatars from a single photograph. Our approach models person appearance by decomposing it into two layers. The first layer is a pose-dependent coarse image that is synthesized by a small neural network. The second layer is defined by a pose-independent texture image that contains high-frequency details and is generated offline. During test-time it is warped and added to the coarse image to ensure high effective resolution of synthesized head views. We compare our system to analogous state-of-the-art systems in terms of visual quality and speed. The experiments show up to an order of magnitude inference speedup over previous neural head avatar models, while achieving state-of-the-art quality. We also report on a real-time smartphone-based implementation of our system.


\clearpage
%
%
\bibliographystyle{splncs04}
\bibliography{refs}

\clearpage
\appendix

\def\x{\mathbf{x}}
\def\y{\mathbf{y}}
\def\w{\mathbf{\omega}}
\def\m{\mathbf{m}}
\def\s{\mathbf{s}}
\def\e{\hat{\mathbf{e}}}
\def\f{\mathbf{f}}

\section{Methods}

We start by explaining training process of our method in much more details. Then, we describe the architecture that we use and how different choices affect the final performance. Finally, we provide a more extended explanation of the mobile inference pipeline that we have adopted.

\subsection{Training details}

We optimize all networks using Adam~\cite{Diederik14} with a learning rate equal to $2\cdot10^{-4}$ $\beta_1 = 0.5$ and $\beta_2 = 0.999$. Before testing, we calculate ``standing'' statistics for all batch normalization layers using 500 mini-batches. Below we provide additional details for the losses that we use.

\subsubsection{Texture mapping regularization.} Below we provide additional implementation details as well as better describe the reasons why this loss is used.

The training signal that the texture generator $G_\text{tex}$ receives is first warped by the warping field $\w^i(t)$ predicted by the inference generator. Because of this, random initializations of the networks typically lead to subpotimal textures, in which the face of the source person occupies a small fraction of the total area of the texture. As the training progresses, this leads to a lower effective resolution of the output image, since the optimization process is unable to escape this bad local optima.

In practice, we address the problem by treating the network's output as a delta to an identity mapping, and also by applying a magnitude penalty on that delta in the early iterations. As mentioned in the main paper, the weight of this penalty is multiplicatively reduced to zero during training, so it does not affect the final performance of the model. More formally, we decompose the output warping field into a sum of two terms: $\w^i(t) = \mathcal{I} + \Delta\w^i(t)$, where $\mathcal{I}$ denotes an identity mapping, and apply an $L_1$ penalty, averaged by a number of spatial positions in the mapping, to the second term:

\begin{equation}
    \mathcal{L}^G_\text{reg} = \frac{1}{HW} ||\Delta\w^i(t)||_1 \, .
\end{equation}

To understand why this regularization helps, we need to briefly describe the implicit properties of the VoxCeleb2 dataset. Since it was obtained using a face detector, a weak from of face alignment is present in the training images, with face occupying more or less the same region.

On the other hand, our regularization allows the gradients to initially flow unperturbed into the texture generator. Therefore, gradients with respect to the texture, averaged over the minibatch, consistently force the texture to produce a high-frequency component of a mean face in the minibatch. This allows the face in the texture to fill the same area as it does in the training images, leading to better generalization.

\subsubsection{Adversarial loss.}

Below we elaborate in more details on the type of adversarial loss that is used. We use the terms~(\ref{eq:score_real})~and~(\ref{eq:score_fake}) to calculate realism scores for real and fake images respectively, with $i_n$ and $t_n$ denoting indices of mini-batch elements, N -- a mini-batch size and $i \in \{ i_1, \dots, i_n \}$:

\begin{equation}\label{eq:score_real}
    \s^i(t) = D \big( \x^i(t), \y^i(t) \big) - \frac{1}{N}\sum_n^N D \big( \hat\x^{i_n}(t_n), \y^{i_n}(t_n) \big) \, ,
\end{equation}

\begin{equation}\label{eq:score_fake}
    \hat\s^i(t) = D \big( \hat\x^i(t), \y^i(t) \big) - \frac{1}{N}\sum_n^N D \big( \x^{i_n}(t_n), \y^{i_n}(t_n) \big) \, .
\end{equation}

Moreover, we use PatchGAN~\cite{Isola17} formulation of the adversarial learning. In it, the discriminator outputs a matrix of realism scores instead of a single prediction, and each element of this matrix is treated as a realism score for a corresponding patch in the input image. This formulation is also used in a large body of relevant works~\cite{Ha19, Wang18b, Wang19} and improves the stability of the adversarial training. If we denote the size of a scores matrix $\s^i(t)$ as $H_s \times W_s$, the resulting objectives can be written as follows:

\begin{equation}\label{eq:loss_adv_dis}
    \mathcal{L}_\text{adv}^D = \frac{1}{H_s W_s} \sum_{h,w} \max \big( 0, 1 - \s_{h,w}^i(t) \big) + \max \big( 0, 1 + \hat\s_{h,w}^i(t) \big) \, ,
\end{equation}

\begin{equation}\label{eq:loss_adv_gen}
    \mathcal{L}_\text{adv}^G = \frac{1}{H_s W_s} \sum_{h,w} \max \big( 0, 1 + \s_{h,w}^i(t) \big) + \max \big( 0, 1 - \hat\s_{h,w}^i(t) \big) \, .
\end{equation}

The loss (\ref{eq:loss_adv_dis}) serves as the discriminator objective. For the generator, we also calculate the feature matching loss \cite{Wang18b}, which has now become a standard component of supervised image-to-image translation models. In this objective, we minimize the distance between the intermediate feature maps of discriminator, calculated using corresponding target and generated images. If we denote as $\f_{k,D}^i(t)$ the features at different spatial resolutions $H_k \times W_k$, then the feature mathing objective is computed as follows:

\begin{equation}\label{eq:loss_fm}
    \mathcal{L}_\text{FM}^G = \frac{1}{K} \sum_k \frac{1}{H_k W_k} || \hat\f_{k,D}^i (t) - \f_{k,D}^i (t) ||_1 \, .
\end{equation}

\subsection{Architecture description}

\begin{figure}[t]
    \centering
    \includegraphics[width=\linewidth]{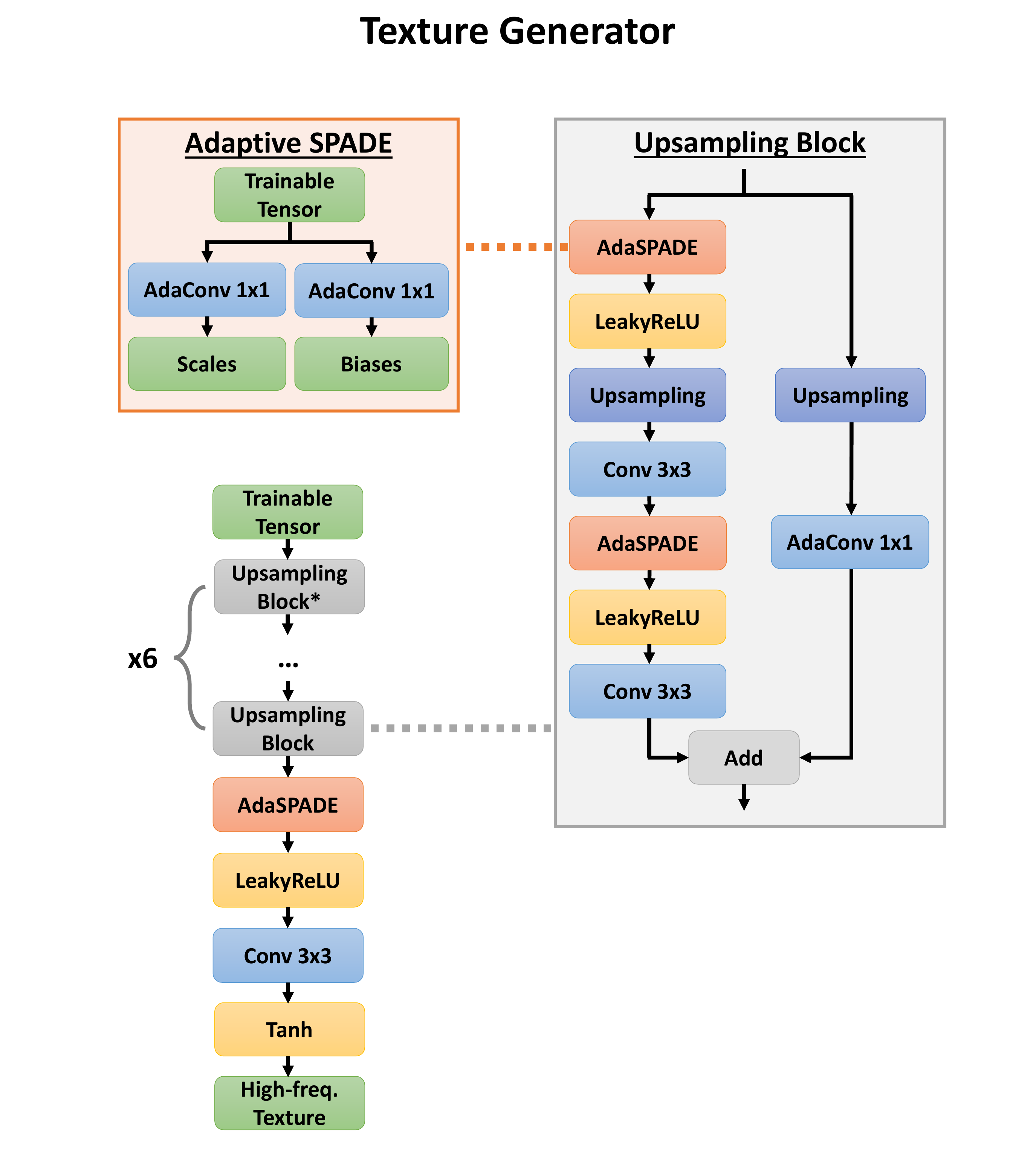}
    \caption{Description of the texture generator's architecture. The first normalization layer in the first upsampling block (marked with a star) is replaced by a regular batch normalization. For the spatial resolution increase, nearest upsampling is performed. All trainable tensors in adaptive SPADE layers have the same size as an output of the previous layer. The first trainable tensor, which is a network's input, has a spatial resolution of $4 \times 4$.}\label{fig:texture_generator_scheme}
\end{figure}

\begin{figure}[t]
    \centering
    \includegraphics[width=\linewidth]{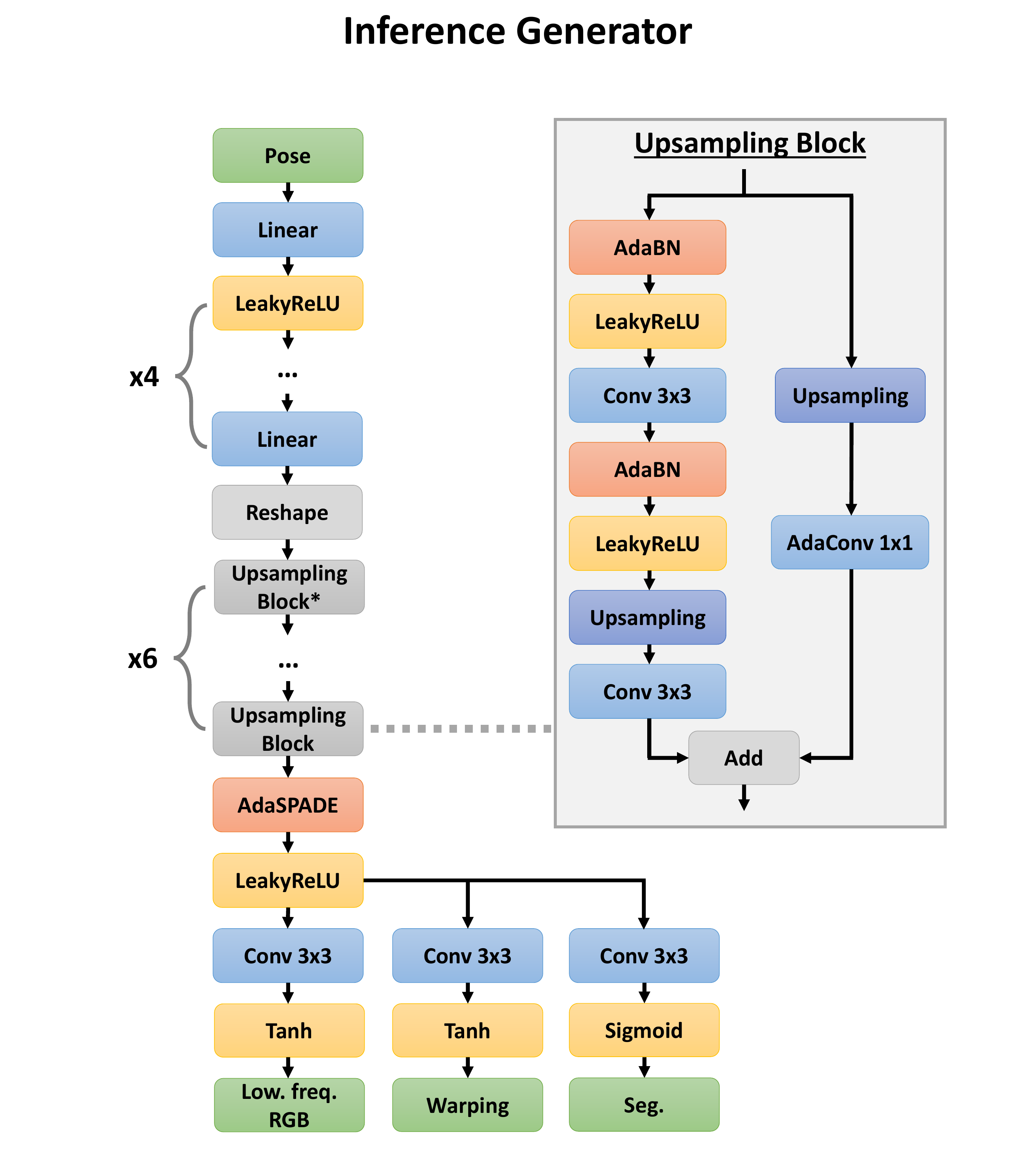}
    \caption{The architecture of the inference generator. As in the texture generator, in the first upsampling block the first normalization layer is replaced by a regular batch normalization. Similarly, nearest upsamping is used. Input pose is reshaped into a vector and fed into a stack of linear layers. Then, the output of the last linear layer is reshaped to have a spatial resolution of $4 \times 4$.}\label{fig:inference_generator_scheme}
\end{figure}

All our networks consist of pre-activation residual blocks. The layout is visualized in the Figures~\ref{fig:texture_generator_scheme}-\ref{fig:updater_scheme}. In all networks, except for the inference generator at the updater, we set the minimum number of channels to 64, and increase (decrease) it by a factor of two each time we perform upsampling (downsampling). We pick the first convolution in each block to increase (decrease) the number of channels. The maximum number of channels is set to 512. In the inference generator we set the minimum number of channels to 32, and the maximum to 256. Also, all linear layers (except for the last one) have their dimensionality set to 256. Moreover, as described in Figure~\ref{fig:inference_generator_scheme}, in the inference generator we employ more efficient blocks, with upsampling performed after the first convolution, and not before it. This allows us to halve the number of MACs per inference.

\begin{figure}[t]
    \centering
    \includegraphics[width=\linewidth]{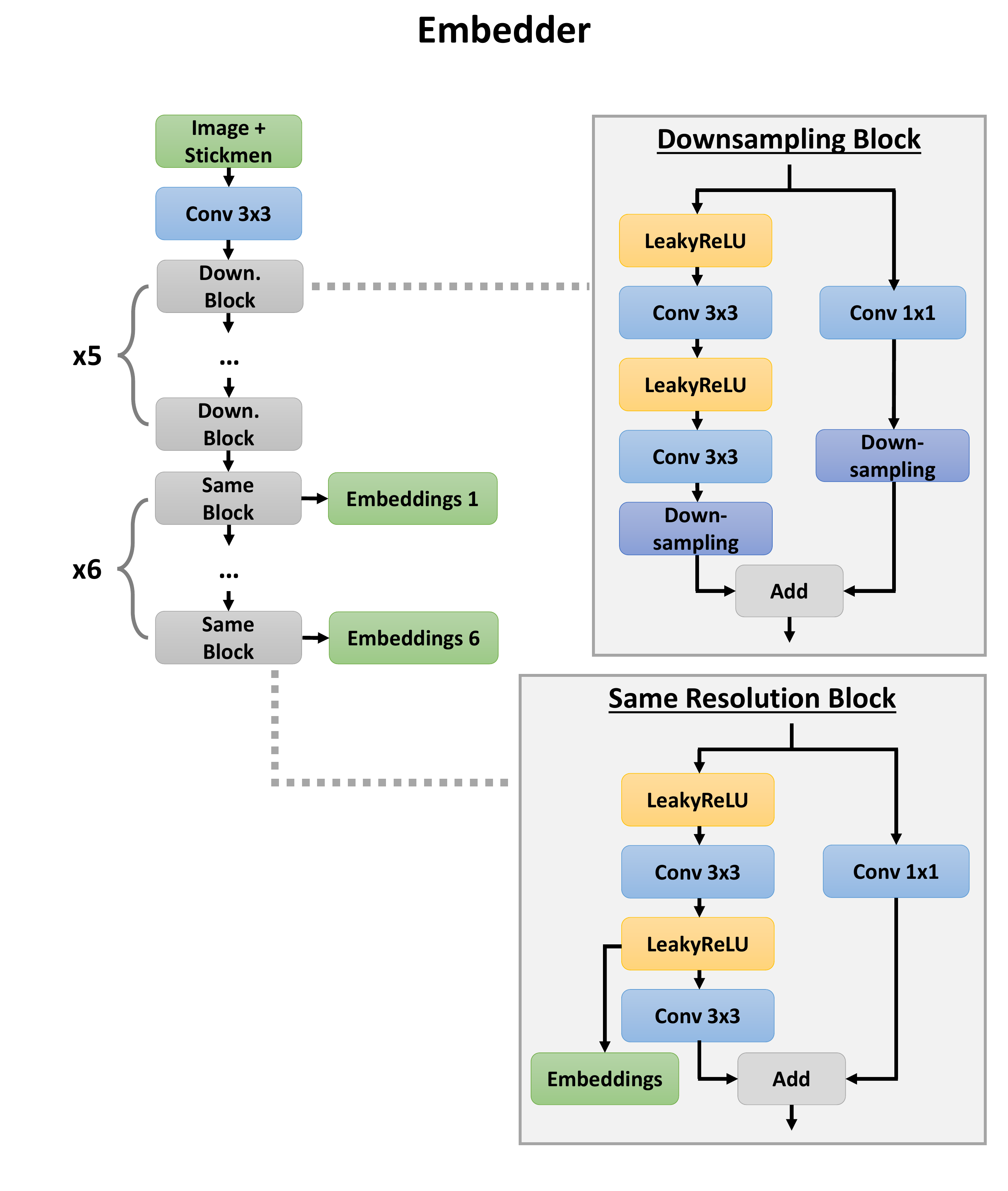}
    \caption{Architecture for the embedder. Here we do not use normalization layers. First, we downsample input images and stickmen to $8 \times 8$ resolution. After that, we obtain embeddings for each of the blocks in the texture and the inference generators. Each embedding is a feature map, and has the same number of channels as the corresponding block in the texture generator. Therefore, we reduce the number of channels in the final blocks, from the maximum of 512 to the minimum of 64 at the end. In the blocks operating at the same resolution, we insert a convolution into a skip connection only when the input and the output number of channels is different.}\label{fig:embedder_scheme}
\end{figure}

In the embedder network (Figure~\ref{fig:embedder_scheme}) each block operating at the same resolution reduces the number of channels, similarly to what is done in the generators. In fact, the output number of channels in each block is excatly equal to the input number of channels in the corresponding generator block. We borrowed this scheme from~\cite{Wang19}, and assume that is it done to botteleneck the embedding tensors, which will be used for the prediction of the adaptive parameters at high resolution. This forces the generators to use all their capacity to generate the image bottom-up, instead of using a shortcut between the source and the target at high resolution, which is present in the architecture.

We do not use batch normalization in the embedder network, because we want it to be trained more slowly, compared to other networks. Otherwise, the whole system overfits to the dataset and the textures become correlated with the source image in terms of head pose. We believe that this is related to the VoxCeleb2 dataset, since in it there is a strong correlation in terms of pose between the randomly sampled source and target frames. This implies that the dataset is lacking diversity with respect to the head movement, and we believe that our system would perform much better either with a better disentangling mechanism of head pose and identity, which we did not come up with, or with a more diverse dataset.

\begin{figure}[t]
    \centering
    \includegraphics[width=\linewidth]{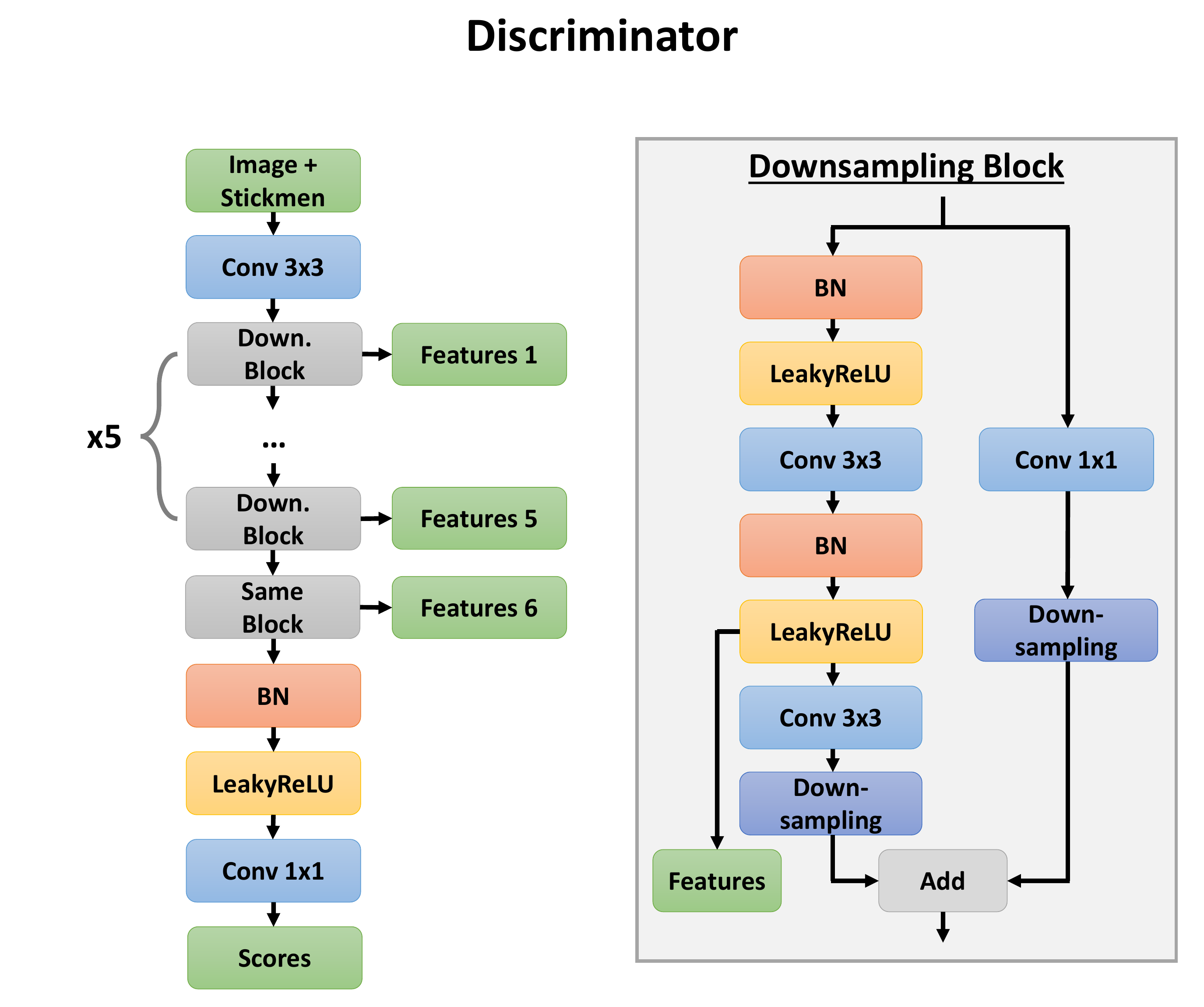}
    \caption{Architecture of the discriminator. We use 5 downsampling blocks and one block operating at final $8 \times 8$ resolution. Additionally, in each block we output features after the second nonlinearity. These features are later used in the feature matching loss. For downsampling, we use average pooling. The architecture of the final block, operating at the same resolution, is similar to the one in the embedder: it is without a convolution in the skip connection, but with batch normalization layers.}\label{fig:discriminator_scheme}
\end{figure}

On contrary, we find it highly beneficial to use batch normalization in the discriminator (Figure~\ref{fig:discriminator_scheme}). This is less memory efficient, compared to the classical scheme, since ``real'' and ``fake'' batches have to be concatenated and fed into the discriminator together. We concatenate these batches to ensure that the first and second order statistics inside the discriminator's features are not whitened with respect to the label (``real'' or ``fake''), which significantly improves the quality of the outputs.

We also tried using instance normalization, but found this to be more sensitive to hyperparameters. For example, the config working on a high-quality dataset cannot be transferred to the low-quality dataset without the occurring instabilities during the adversarial training.

We predict adaptive parameters following the procedure inspired by a matrix decomposition. The basic idea is to predict a weight tensor for the convolution via a decomposition of the embedding tensor. In our work, we use the following procedure (taken from~\cite{Wang19}) to predict the weights for all $1 \times 1$ convolutions and adaptive batch normalization layers in the texture and the inference generators:

\begin{itemize}
    \item Resize all embedding tensors $\e^i_k(s)$, with the number of channels $C_k$, by nearest upsampling to $32 \times 32$ resolution for the texture generator, and $16 \times 16$ for the medium-sized inference generator. \\
    \item Flatten the resized tensor across its spatial resoluton, converting it to a matrix of the shape $C_k \times 1024$ for the texture generator, and $\frac{1}{2} C_k \times 512$ for the inference generator (the first dimensionality has to match the reduced number of channels in the convolutions of the medium-sized model). \\
    \item Three linear layers (with no nonlinearities in between) are then applied, performing the decomposition. A resulting matrix should match the shape of the weights, combined with the biases, for each specific adaptive layer. These linear layers are trained separately for each adaptive convolution and adaptive batch normalization.
\end{itemize}

Each embedding tensor $\e^i_k(s)$ is therefore used to predict all adaptive parameters inside the layers of the $k$-th block in the texture and inference generators. We do not perform an ablation study with respect to this scheme, since it was used in an already published work on a similar topic.

\begin{figure}[t]
    \centering
    \includegraphics[width=\linewidth]{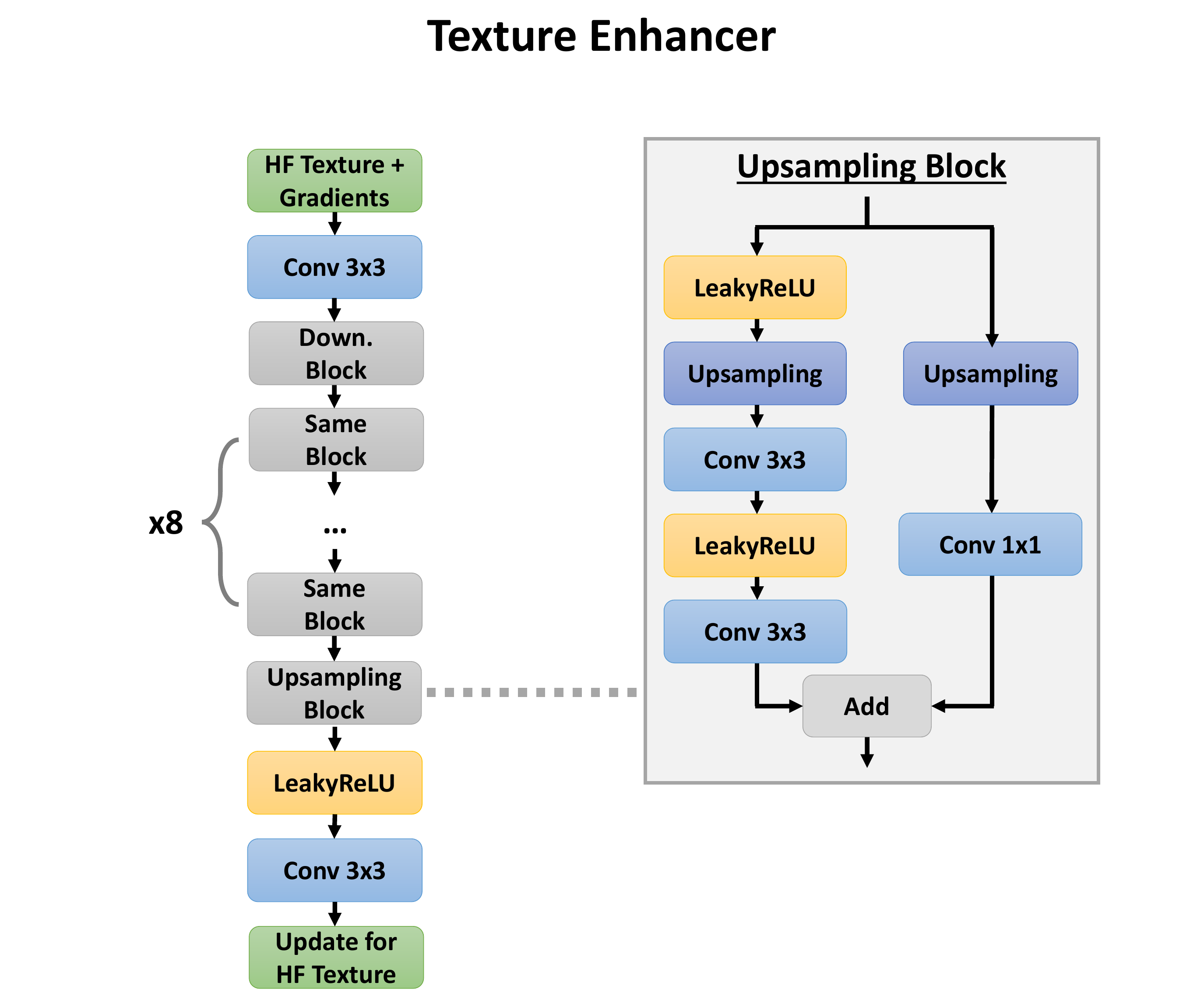}
    \caption{We employ a simple encoder-decoder style architecture, similar to the one used in~\cite{Isola17}. We replace downsampling and upsampling layers with residual blocks. We also do not employ batch normalization inside the enhancer.}\label{fig:updater_scheme}
\end{figure}

Finally, we describe the architecture of the texture enhancer in Figure~\ref{fig:updater_scheme}. This architecture is standard for image-to-image translation tasks. The spatial dimensionality and the number of channels in the bottleneck is equal to 128.

\subsection{Mobile inference}

As mentioned in main paper, we train our models using PyTorch and then port them to smartphones with Qualcomm Snapdragon 855 chips. For inference, we use a native Snapdragon Neural Processing Engine (SNPE) APK, which provides a significant speed-up compared to TF-Lite and PyTorch mobile. In order to convert the models trained in PyTorch into SNPE-compatible containers, we first use the PyTorch-ONNX parser, as it is simple to get an ONNX model right from PyTorch. However, it does not guarantee that the obtained model can be converted into a mobile-compatible container, since some operations may be unsupported by SNPE. Moreover, there is a collision between different versions of ONNX and SNPE operation sets, with some versions of the operations being incompatible with each other. We have solved this problem by using PyTorch 1.3 and SNPE 1.32, but solely for operations used our inference generator. This is part of the reason why we had to resort to simple layers, like BathNorm-s, convolutions and nonlinearities in our network..

All ported models have spectral normalization removed, and adaptive parameters fixed and merged into their base layers. In our experiments the target platform is Adreno 640 GPU, utilized in FP16 mode. We do not observe any noticeable quality degradation from running our model in FP16 (although training in FP16 or mixed precision settings leads to instabilities and early explosion of the gradients). Since our model includes bilinear sampling from texture (using a predicted warping field), that is not supported by SNPE, we implement it ourselves, as a part of application, called after each inferred frame on a CPU. The GPU implementation should be possible as well, but is more time-consuming to implement. Our reported mobile timings (42 ms, averaged by 100 runs) do not include the bilinear sampling and copy operations from GPU to CPU. On CPU, bilinear sampling takes additional 2 milliseconds, but for a GPU implementation, the timing would be negligible.

\section{Experiments}

\subsection{Training details for the state-of-the-art methods.}

First Order Motion model was trained using a config provided with the official implementation of the model. In order to obtain a family of models, we modify minimum and maximum number of channels in the generator from default 64 and 512 to 32 and 256 for the medium, and 16 and 128 for the small models.

For Few-shot Vid-to-Vid, we have also used a default config from the official implementation, but with slight modifications. Since we train on a dataset with videos already being cropped, we removed the random crop and scale augmentations in order to avoid a domain gap between training and testing. In our case, that would lead to black borders appearing on the training images, and a suboptimal performance on a test set with no such artifacts. In order to obtain a family of models, we also reduce the minimum and maximum number of channels in the generator from the default 32 and 1024 to 32 and 256 for the medium model and 16 and 128 for the small model.

To calculate the number of multiply-accumulate operations, we used an off-the-shelf tool that evaluates this number for all internal PyTorch modules. That way of calculation, while being easy, is not perfect as, for example, it does not account for the number of operations in PyTorch functionals, which may be called inside the model. Other forms of complexity evaluation would require significant refactor of the code of the competitors, which lies out of the scope of our comparison. For our model, we have provided accurate complexity esimates.

\subsection{Extended evaluations.}

\begin{figure}[t]
    \centering    
    \setlength{\wid}{0.155\textwidth}
    \begin{tabular}{cccccc}
        \includegraphics[align=c,bmargin=0.1cm,width=\wid]{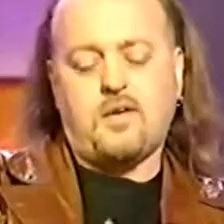} &
        \includegraphics[align=c,width=\wid]{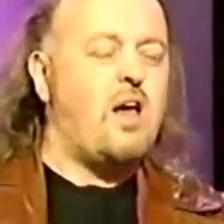} &
        \includegraphics[align=c,width=\wid]{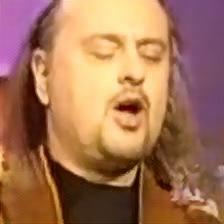} &
        \includegraphics[align=c,width=\wid]{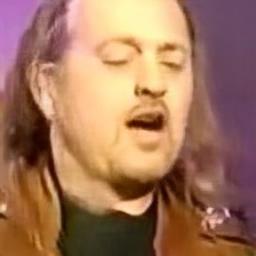} &
        \includegraphics[align=c,width=\wid]{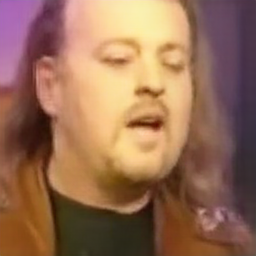} &
        \includegraphics[align=c,width=\wid]{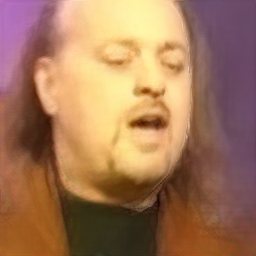} \\
        \includegraphics[align=c,bmargin=0.1cm,width=\wid]{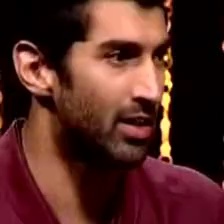} &
        \includegraphics[align=c,width=\wid]{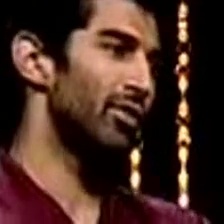} &
        \includegraphics[align=c,width=\wid]{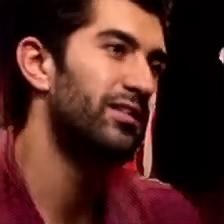} &
        \includegraphics[align=c,width=\wid]{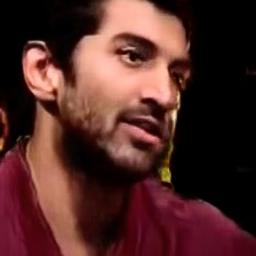} &
        \includegraphics[align=c,width=\wid]{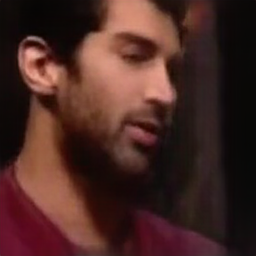} &
        \includegraphics[align=c,width=\wid]{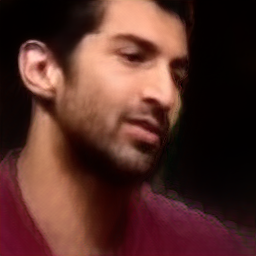} \\
        \includegraphics[align=c,bmargin=0.1cm,width=\wid]{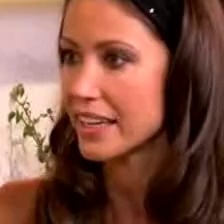} &
        \includegraphics[align=c,width=\wid]{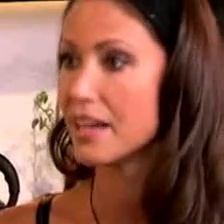} &
        \includegraphics[align=c,width=\wid]{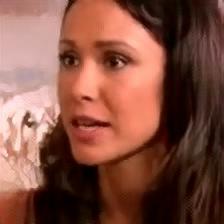} &
        \includegraphics[align=c,width=\wid]{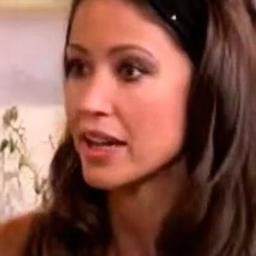} &
        \includegraphics[align=c,width=\wid]{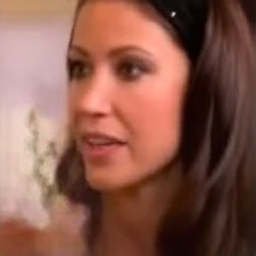} &
        \includegraphics[align=c,width=\wid]{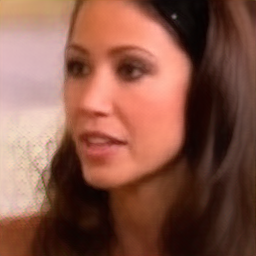} \\
        \includegraphics[align=c,bmargin=0.1cm,width=\wid]{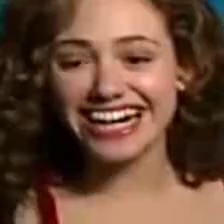} &
        \includegraphics[align=c,width=\wid]{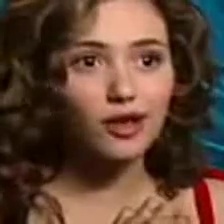} &
        \includegraphics[align=c,width=\wid]{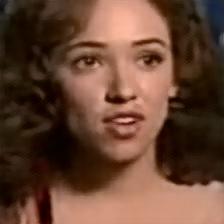} &
        \includegraphics[align=c,width=\wid]{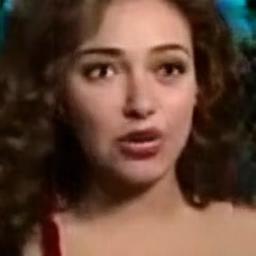} &
        \includegraphics[align=c,width=\wid]{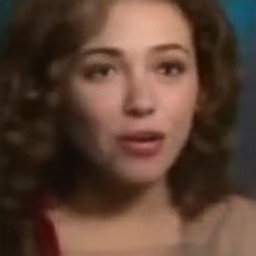} &
        \includegraphics[align=c,width=\wid]{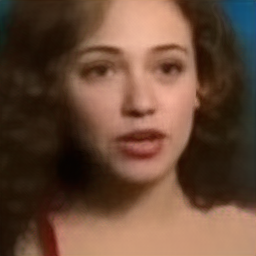} \\
        \includegraphics[align=c,bmargin=0.1cm,width=\wid]{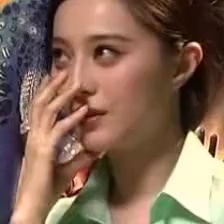} &
        \includegraphics[align=c,width=\wid]{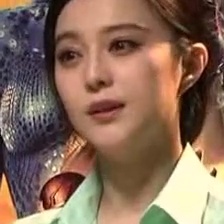} &
        \includegraphics[align=c,width=\wid]{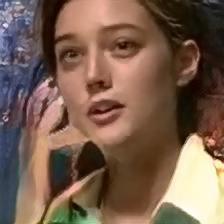} &
        \includegraphics[align=c,width=\wid]{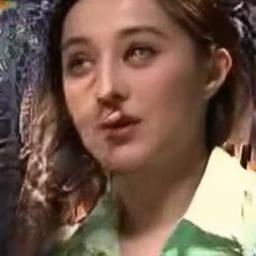} &
        \includegraphics[align=c,width=\wid]{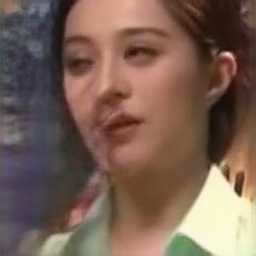} &
        \includegraphics[align=c,width=\wid]{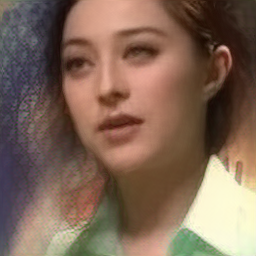} \\
        \includegraphics[align=c,bmargin=0.1cm,width=\wid]{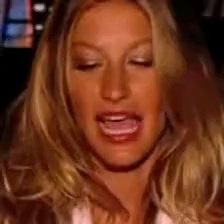} &
        \includegraphics[align=c,width=\wid]{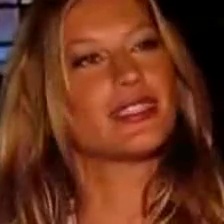} &
        \includegraphics[align=c,width=\wid]{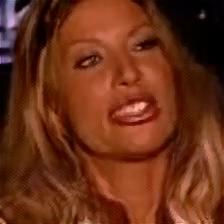} &
        \includegraphics[align=c,width=\wid]{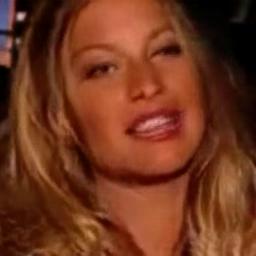} &
        \includegraphics[align=c,width=\wid]{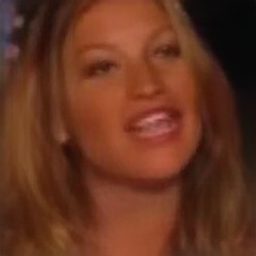} &
        \includegraphics[align=c,width=\wid]{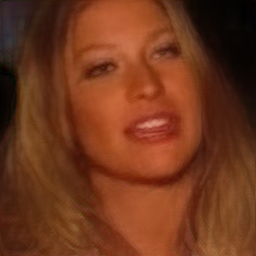} \\
        \includegraphics[align=c,bmargin=0.1cm,width=\wid]{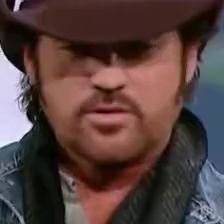} &
        \includegraphics[align=c,width=\wid]{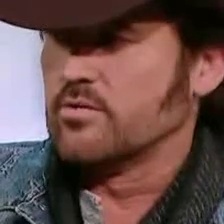} &
        \includegraphics[align=c,width=\wid]{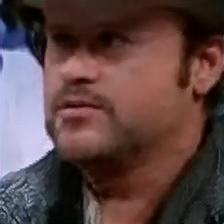} &
        \includegraphics[align=c,width=\wid]{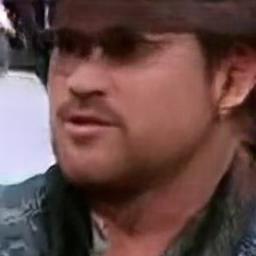} &
        \includegraphics[align=c,width=\wid]{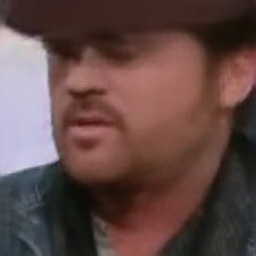} &
        \includegraphics[align=c,width=\wid]{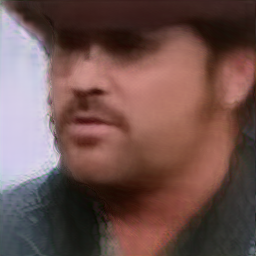} \\
        \includegraphics[align=c,bmargin=0.1cm,width=\wid]{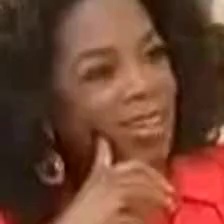} &
        \includegraphics[align=c,width=\wid]{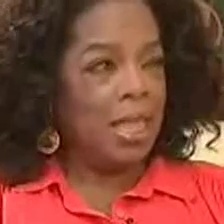} &
        \includegraphics[align=c,width=\wid]{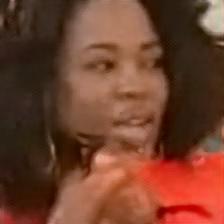} &
        \includegraphics[align=c,width=\wid]{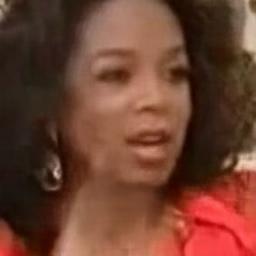} &
        \includegraphics[align=c,width=\wid]{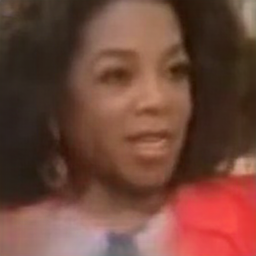} &
        \includegraphics[align=c,width=\wid]{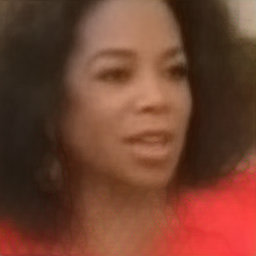} \\
        \textbf{Source} & \textbf{Target} & \textbf{\begin{tabular}{c} Few-shot \\ T. Heads \end{tabular}} & \textbf{\begin{tabular}{c} Few-shot \\ Vid-to-Vid \end{tabular}} & \textbf{FOMM} & \textbf{Ours}
    \end{tabular}
    \caption{Extended comparison of the medium-sized models from all method families on the VoxCeleb2 dataset. For Few-shot Talking Heads we use the results obtained using the original full-sized model.}
    \label{fig:ext_vc2_comp}
\end{figure}

\begin{figure}[t]
    \centering    
    \setlength{\wid}{0.155\textwidth}
    \begin{tabular}{cccccc}
        \includegraphics[align=c,width=\wid,bmargin=0.1cm]{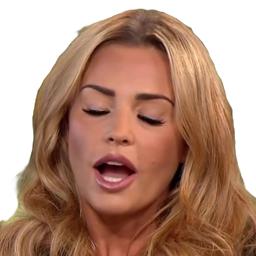} &
        \includegraphics[align=c,width=\wid,bmargin=0.1cm]{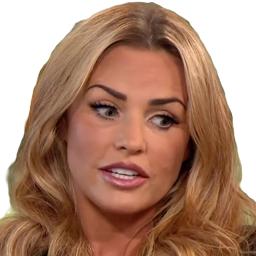} &
        \includegraphics[align=c,width=\wid,bmargin=0.1cm]{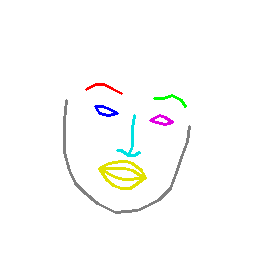} &
        \includegraphics[align=c,width=\wid,bmargin=0.1cm]{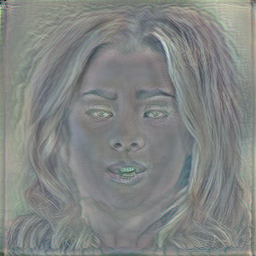} & 
        \includegraphics[align=c,width=\wid,bmargin=0.1cm]{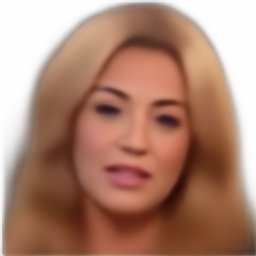} &
        \includegraphics[align=c,width=\wid,bmargin=0.1cm]{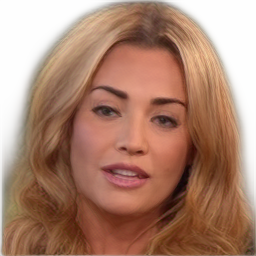} \\
        \includegraphics[align=c,width=\wid,bmargin=0.1cm]{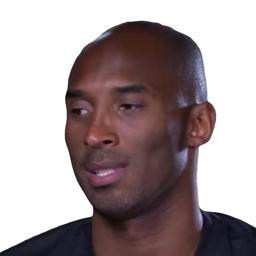} &
        \includegraphics[align=c,width=\wid,bmargin=0.1cm]{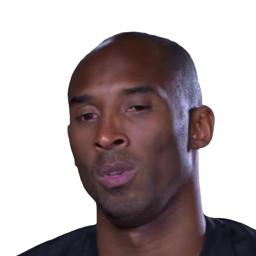} &
        \includegraphics[align=c,width=\wid,bmargin=0.1cm]{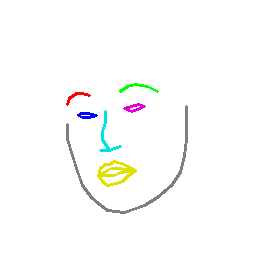} &
        \includegraphics[align=c,width=\wid,bmargin=0.1cm]{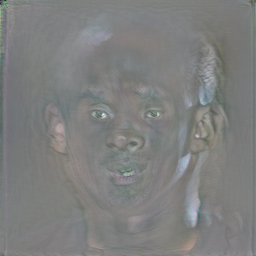} &
        \includegraphics[align=c,width=\wid,bmargin=0.1cm]{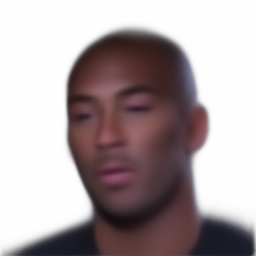} &
        \includegraphics[align=c,width=\wid,bmargin=0.1cm]{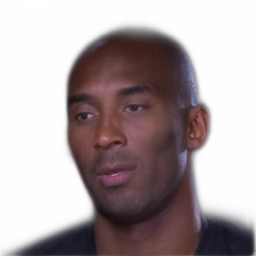} \\ 
        \includegraphics[align=c,width=\wid,bmargin=0.1cm]{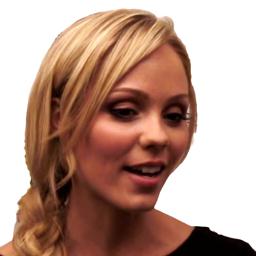} &
        \includegraphics[align=c,width=\wid,bmargin=0.1cm]{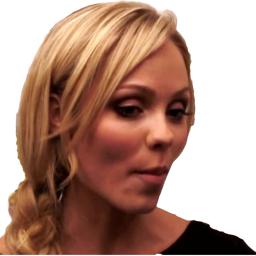} &
        \includegraphics[align=c,width=\wid,bmargin=0.1cm]{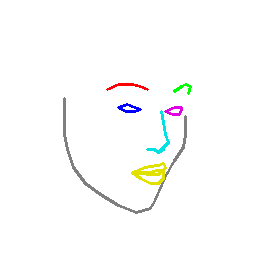} &
        \includegraphics[align=c,width=\wid,bmargin=0.1cm]{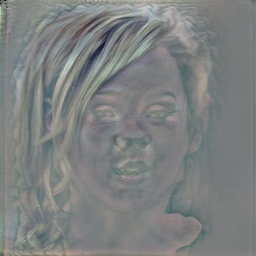} &
        \includegraphics[align=c,width=\wid,bmargin=0.1cm]{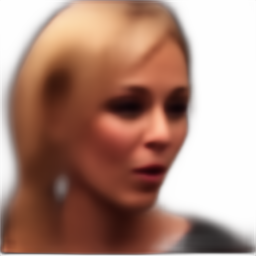} &
        \includegraphics[align=c,width=\wid,bmargin=0.1cm]{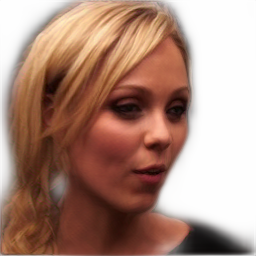} \\
        \includegraphics[align=c,width=\wid,bmargin=0.1cm]{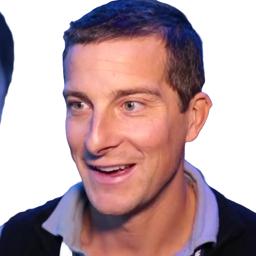} &
        \includegraphics[align=c,width=\wid,bmargin=0.1cm]{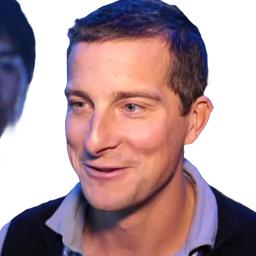} &
        \includegraphics[align=c,width=\wid,bmargin=0.1cm]{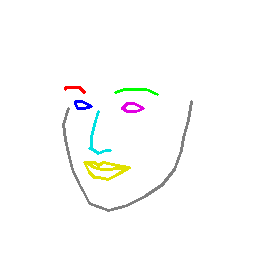} &
        \includegraphics[align=c,width=\wid,bmargin=0.1cm]{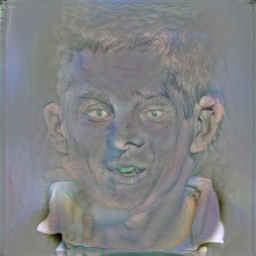} &
        \includegraphics[align=c,width=\wid,bmargin=0.1cm]{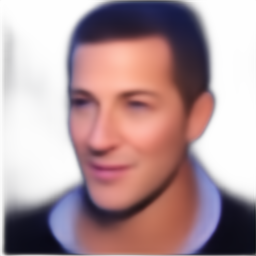} &
        \includegraphics[align=c,width=\wid,bmargin=0.1cm]{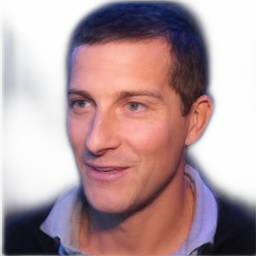} \\
        \includegraphics[align=c,width=\wid,bmargin=0.1cm]{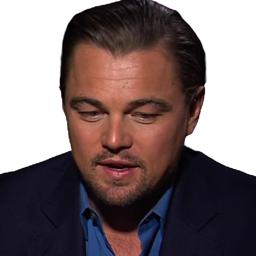} &
        \includegraphics[align=c,width=\wid,bmargin=0.1cm]{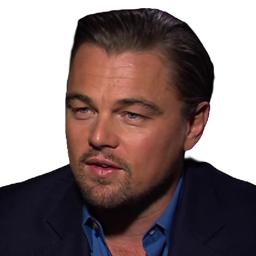} &
        \includegraphics[align=c,width=\wid,bmargin=0.1cm]{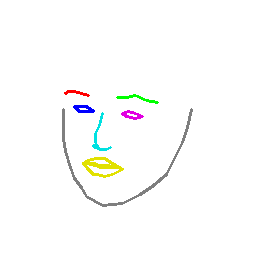} &
        \includegraphics[align=c,width=\wid,bmargin=0.1cm]{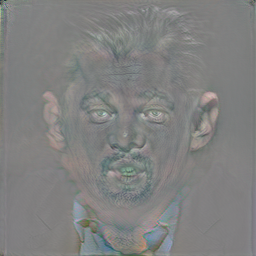} & 
        \includegraphics[align=c,width=\wid,bmargin=0.1cm]{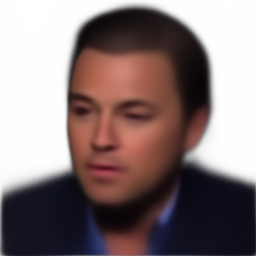} &
        \includegraphics[align=c,width=\wid,bmargin=0.1cm]{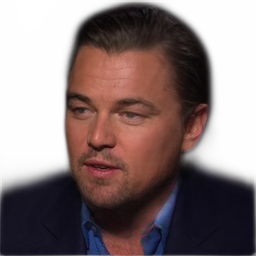} \\
        \includegraphics[align=c,width=\wid,bmargin=0.1cm]{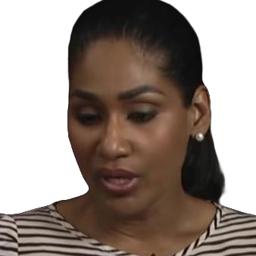} &
        \includegraphics[align=c,width=\wid,bmargin=0.1cm]{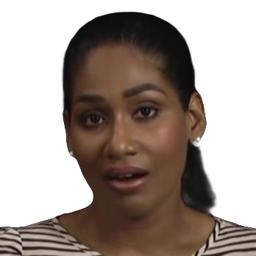} &
        \includegraphics[align=c,width=\wid,bmargin=0.1cm]{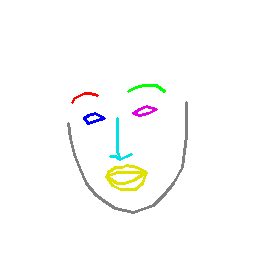} &
        \includegraphics[align=c,width=\wid,bmargin=0.1cm]{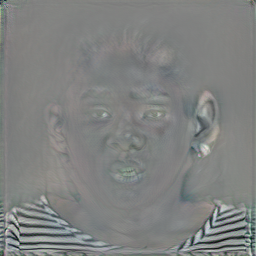} &
        \includegraphics[align=c,width=\wid,bmargin=0.1cm]{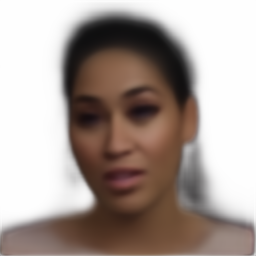} &
        \includegraphics[align=c,width=\wid,bmargin=0.1cm]{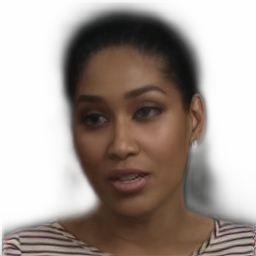} \\ 
        \includegraphics[align=c,width=\wid,bmargin=0.1cm]{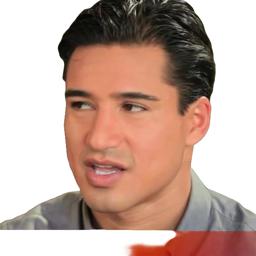} &
        \includegraphics[align=c,width=\wid,bmargin=0.1cm]{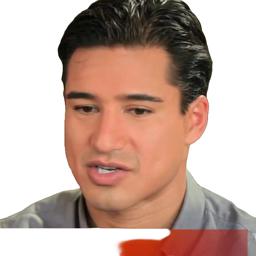} &
        \includegraphics[align=c,width=\wid,bmargin=0.1cm]{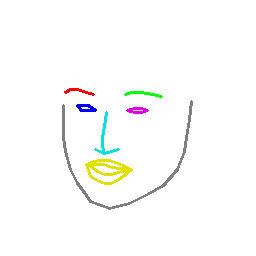} &
        \includegraphics[align=c,width=\wid,bmargin=0.1cm]{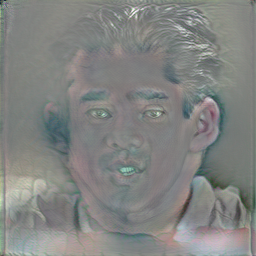} &
        \includegraphics[align=c,width=\wid,bmargin=0.1cm]{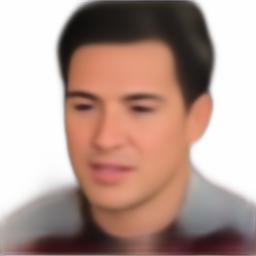} &
        \includegraphics[align=c,width=\wid,bmargin=0.1cm]{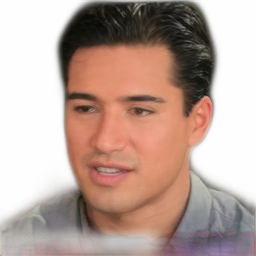} \\
        \includegraphics[align=c,width=\wid,bmargin=0.1cm]{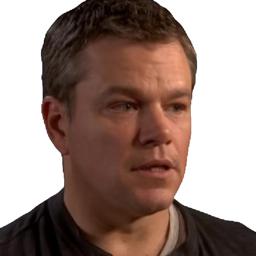} &
        \includegraphics[align=c,width=\wid,bmargin=0.1cm]{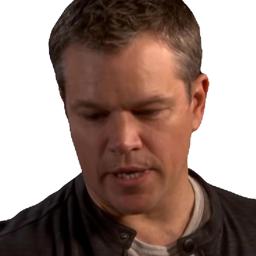} &
        \includegraphics[align=c,width=\wid,bmargin=0.1cm]{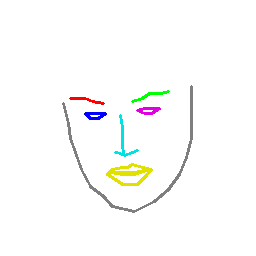} &
        \includegraphics[align=c,width=\wid,bmargin=0.1cm]{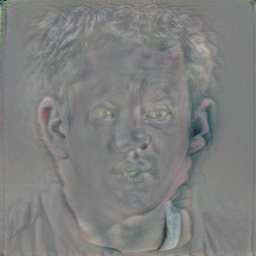} &
        \includegraphics[align=c,width=\wid,bmargin=0.1cm]{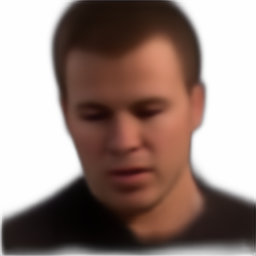} &
        \includegraphics[align=c,width=\wid,bmargin=0.1cm]{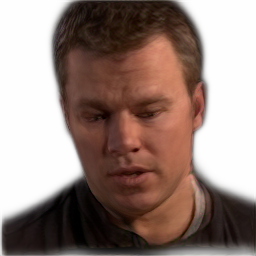} \\
        \textbf{Source} & \textbf{Target} & \textbf{Pose} & \textbf{Texture} & \textbf{Low-freq.} & \textbf{Ours}
    \end{tabular}
    \caption{Detailed qualitative results for our medium-sized model trained on the VoxCeleb2-HQ dataset.}
    \label{fig:ext_detailed_results}
\end{figure}

\begin{figure}[t]
    \centering    
    \setlength{\wid}{0.148\textwidth}
    \begin{tabular}{m{1em}cccccc}
        \rotatebox{90}{\textbf{Small}} &
        \includegraphics[align=c,width=\wid,bmargin=0.1cm]{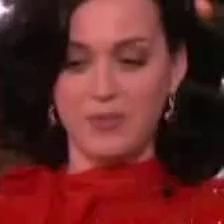} &
        \includegraphics[align=c,width=\wid,bmargin=0.1cm]{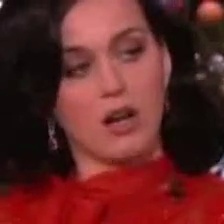} &
        \includegraphics[align=c,width=\wid,bmargin=0.1cm]{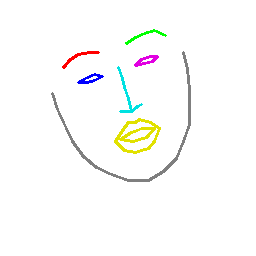} &
        \includegraphics[align=c,width=\wid,bmargin=0.1cm]{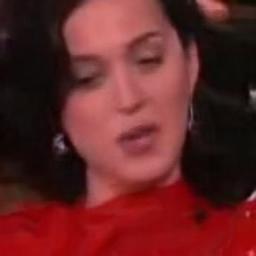} &
        \includegraphics[align=c,width=\wid,bmargin=0.1cm]{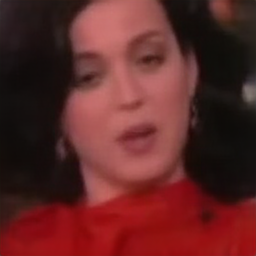} &
        \includegraphics[align=c,width=\wid,bmargin=0.1cm]{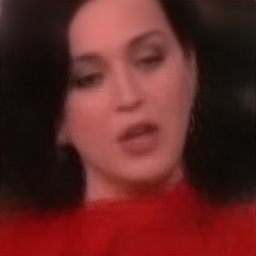} \\
        \rotatebox{90}{\textbf{Medium}} &
        \includegraphics[align=c,width=\wid,bmargin=0.1cm]{figures_suppmat/vc2_family_comp/id04656_source.jpg} &
        \includegraphics[align=c,width=\wid,bmargin=0.1cm]{figures_suppmat/vc2_family_comp/id04656_target.jpg} &
        \includegraphics[align=c,width=\wid,bmargin=0.1cm]{figures_suppmat/vc2_family_comp/id04656_pose.png} &
        \includegraphics[align=c,width=\wid,bmargin=0.1cm]{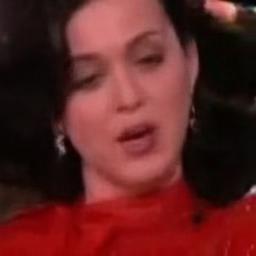} &
        \includegraphics[align=c,width=\wid,bmargin=0.1cm]{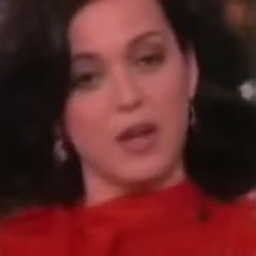} &
        \includegraphics[align=c,width=\wid,bmargin=0.1cm]{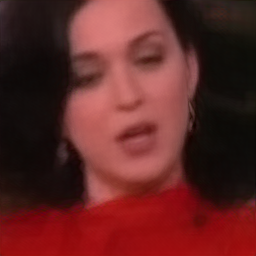} \\
        \rotatebox{90}{\textbf{Large}} &
        \includegraphics[align=c,width=\wid,bmargin=0.1cm]{figures_suppmat/vc2_family_comp/id04656_source.jpg} &
        \includegraphics[align=c,width=\wid,bmargin=0.1cm]{figures_suppmat/vc2_family_comp/id04656_target.jpg} &
        \includegraphics[align=c,width=\wid,bmargin=0.1cm]{figures_suppmat/vc2_family_comp/id04656_pose.png} &
        \includegraphics[align=c,width=\wid,bmargin=0.1cm]{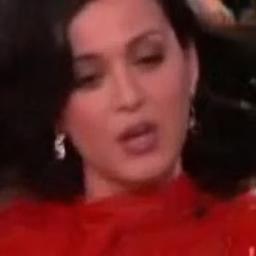} &
        \includegraphics[align=c,width=\wid,bmargin=0.1cm]{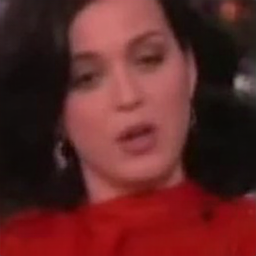} &
        \includegraphics[align=c,width=\wid,bmargin=0.1cm]{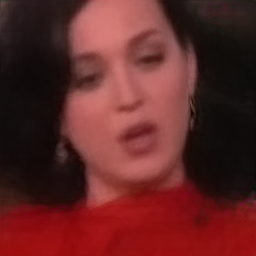} \\
        \rotatebox{90}{\textbf{Small}} &
        \includegraphics[align=c,width=\wid,bmargin=0.1cm]{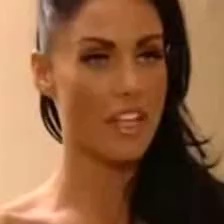} &
        \includegraphics[align=c,width=\wid,bmargin=0.1cm]{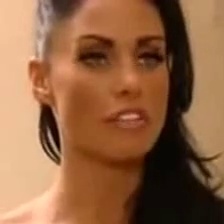} &
        \includegraphics[align=c,width=\wid,bmargin=0.1cm]{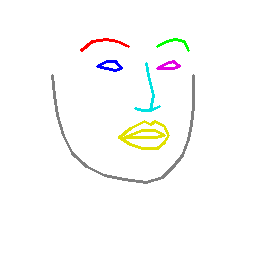} &
        \includegraphics[align=c,width=\wid,bmargin=0.1cm]{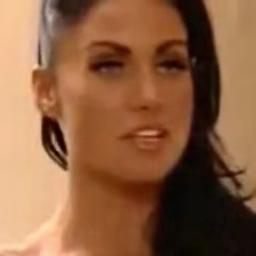} &
        \includegraphics[align=c,width=\wid,bmargin=0.1cm]{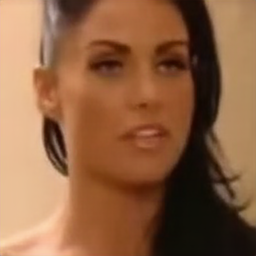} &
        \includegraphics[align=c,width=\wid,bmargin=0.1cm]{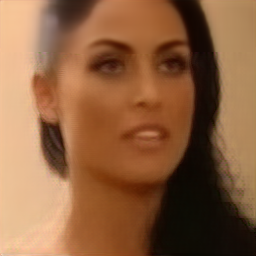} \\
        \rotatebox{90}{\textbf{Medium}} &
        \includegraphics[align=c,width=\wid,bmargin=0.1cm]{figures_suppmat/vc2_family_comp/id04657_source.jpg} &
        \includegraphics[align=c,width=\wid,bmargin=0.1cm]{figures_suppmat/vc2_family_comp/id04657_target.jpg} &
        \includegraphics[align=c,width=\wid,bmargin=0.1cm]{figures_suppmat/vc2_family_comp/id04657_pose.png} &
        \includegraphics[align=c,width=\wid,bmargin=0.1cm]{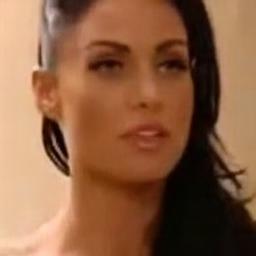} &
        \includegraphics[align=c,width=\wid,bmargin=0.1cm]{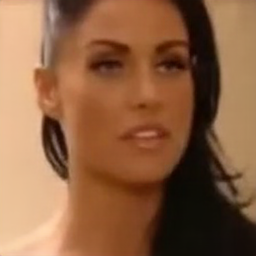} &
        \includegraphics[align=c,width=\wid,bmargin=0.1cm]{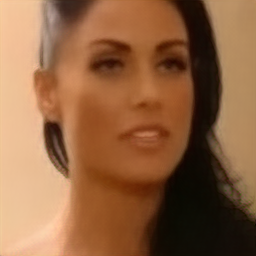} \\
        \rotatebox{90}{\textbf{Large}} &
        \includegraphics[align=c,width=\wid,bmargin=0.1cm]{figures_suppmat/vc2_family_comp/id04657_source.jpg} &
        \includegraphics[align=c,width=\wid,bmargin=0.1cm]{figures_suppmat/vc2_family_comp/id04657_target.jpg} &
        \includegraphics[align=c,width=\wid,bmargin=0.1cm]{figures_suppmat/vc2_family_comp/id04657_pose.png} &
        \includegraphics[align=c,width=\wid,bmargin=0.1cm]{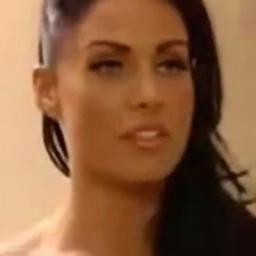} &
        \includegraphics[align=c,width=\wid,bmargin=0.1cm]{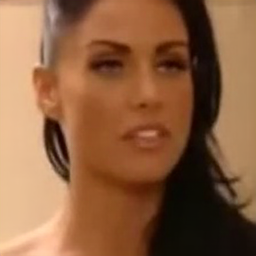} &
        \includegraphics[align=c,width=\wid,bmargin=0.1cm]{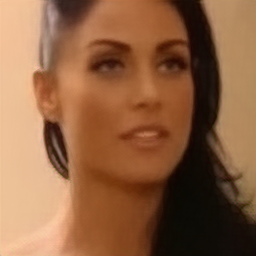} \\
        \rotatebox{90}{\textbf{Small}} &
        \includegraphics[align=c,width=\wid,bmargin=0.1cm]{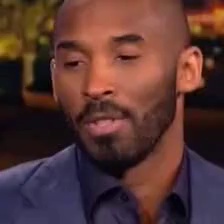} &
        \includegraphics[align=c,width=\wid,bmargin=0.1cm]{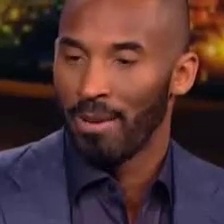} &
        \includegraphics[align=c,width=\wid,bmargin=0.1cm]{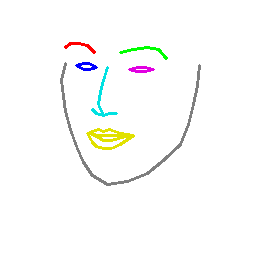} &
        \includegraphics[align=c,width=\wid,bmargin=0.1cm]{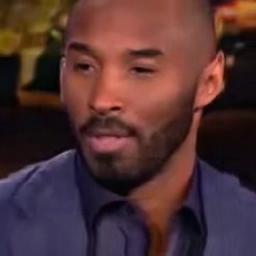} &
        \includegraphics[align=c,width=\wid,bmargin=0.1cm]{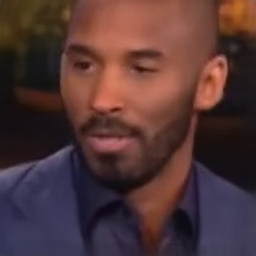} &
        \includegraphics[align=c,width=\wid,bmargin=0.1cm]{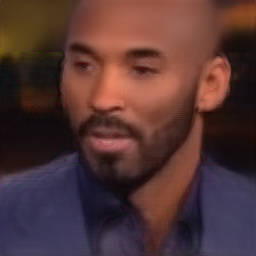} \\
        \rotatebox{90}{\textbf{Medium}} &
        \includegraphics[align=c,width=\wid,bmargin=0.1cm]{figures_suppmat/vc2_family_comp/id04862_source.jpg} &
        \includegraphics[align=c,width=\wid,bmargin=0.1cm]{figures_suppmat/vc2_family_comp/id04862_target.jpg} &
        \includegraphics[align=c,width=\wid,bmargin=0.1cm]{figures_suppmat/vc2_family_comp/id04862_pose.png} &
        \includegraphics[align=c,width=\wid,bmargin=0.1cm]{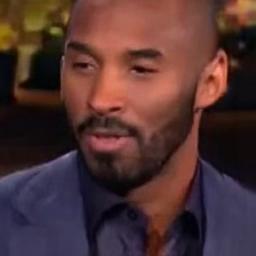} &
        \includegraphics[align=c,width=\wid,bmargin=0.1cm]{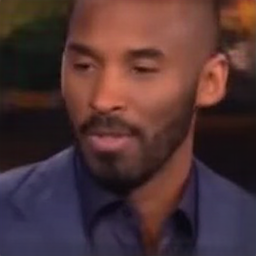} &
        \includegraphics[align=c,width=\wid,bmargin=0.1cm]{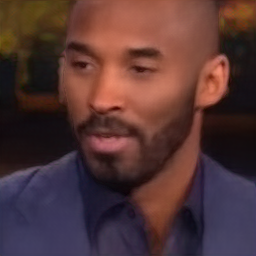} \\
        \rotatebox{90}{\textbf{Large}} &
        \includegraphics[align=c,width=\wid,bmargin=0.1cm]{figures_suppmat/vc2_family_comp/id04862_source.jpg} &
        \includegraphics[align=c,width=\wid,bmargin=0.1cm]{figures_suppmat/vc2_family_comp/id04862_target.jpg} &
        \includegraphics[align=c,width=\wid,bmargin=0.1cm]{figures_suppmat/vc2_family_comp/id04862_pose.png} &
        \includegraphics[align=c,width=\wid,bmargin=0.1cm]{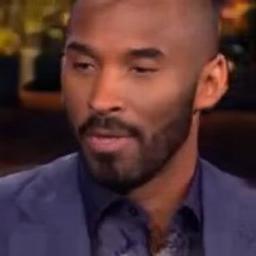} &
        \includegraphics[align=c,width=\wid,bmargin=0.1cm]{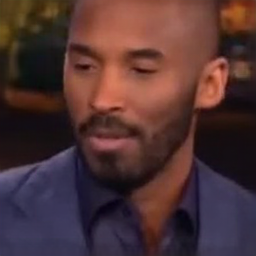} &
        \includegraphics[align=c,width=\wid,bmargin=0.1cm]{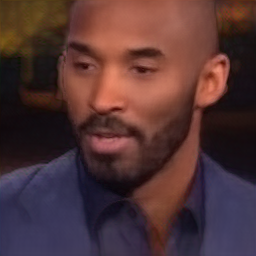}  \\
        & \textbf{Source} & \textbf{Target} & \textbf{Pose} & \textbf{\begin{tabular}{c} Few-shot \\ Vid-to-Vid \end{tabular}} & \textbf{FOMM} & \textbf{Ours}
    \end{tabular} 
    \caption{Qualitative comparison between the small, medium and large models for all compared families of methods. }
    \label{fig:vc2_family_comp}
\end{figure}

\begin{figure}[t]
    \centering    
    \setlength{\wid}{0.155\textwidth}
    \begin{tabular}{cccccc}
        \includegraphics[align=c,width=\wid,bmargin=0.1cm]{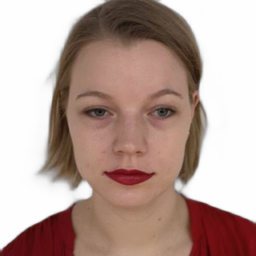} &
        \includegraphics[align=c,width=\wid,bmargin=0.1cm]{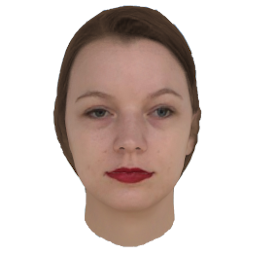} &
        \includegraphics[align=c,width=\wid,bmargin=0.1cm]{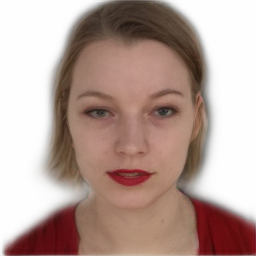} &
        \includegraphics[align=c,width=\wid,bmargin=0.1cm]{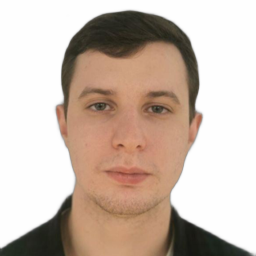} & 
        \includegraphics[align=c,width=\wid,bmargin=0.1cm]{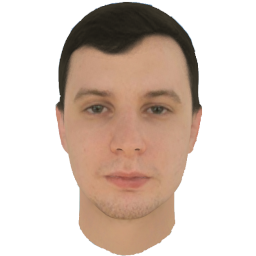} &
        \includegraphics[align=c,width=\wid,bmargin=0.1cm]{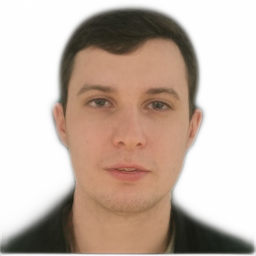} \\
        \includegraphics[align=c,width=\wid,bmargin=0.1cm]{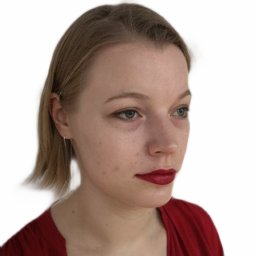} &
        \includegraphics[align=c,width=\wid,bmargin=0.1cm]{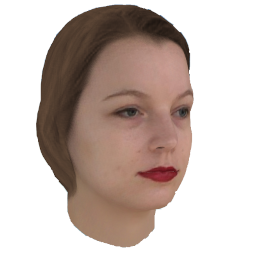} &
        \includegraphics[align=c,width=\wid,bmargin=0.1cm]{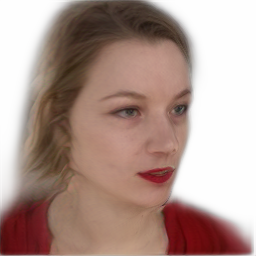} &
        \includegraphics[align=c,width=\wid,bmargin=0.1cm]{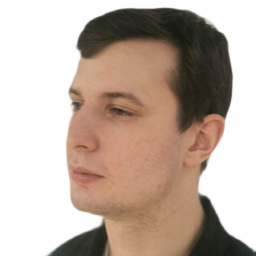} & 
        \includegraphics[align=c,width=\wid,bmargin=0.1cm]{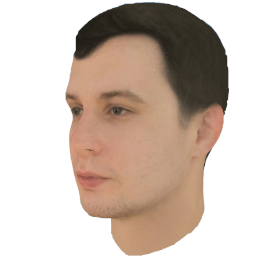} &
        \includegraphics[align=c,width=\wid,bmargin=0.1cm]{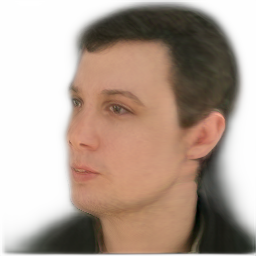} \\
        \includegraphics[align=c,width=\wid,bmargin=0.1cm]{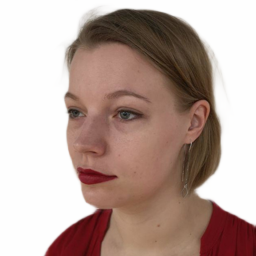} &
        \includegraphics[align=c,width=\wid,bmargin=0.1cm]{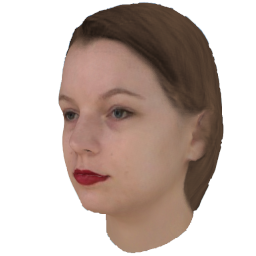} &
        \includegraphics[align=c,width=\wid,bmargin=0.1cm]{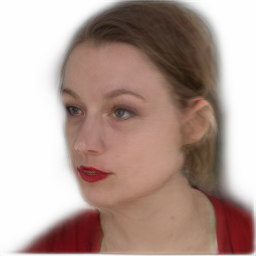} &
        \includegraphics[align=c,width=\wid,bmargin=0.1cm]{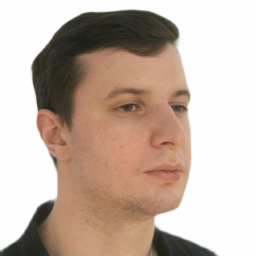} & 
        \includegraphics[align=c,width=\wid,bmargin=0.1cm]{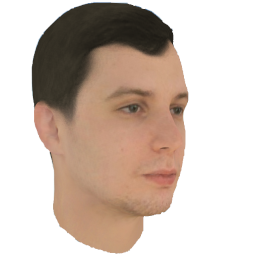} &
        \includegraphics[align=c,width=\wid,bmargin=0.1cm]{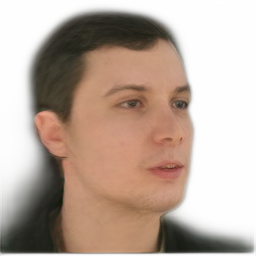} \\
        \textbf{Target} & \textbf{AvatarSDK} & \textbf{Ours} & \textbf{Target} & \textbf{AvatarSDK} & \textbf{Ours}
    \end{tabular}
    \caption{Comparison of our method with a closed-source product~\cite{AvatarSDK}, which is representative of the state-of-the-art in real-time one-shot avatar creation, based on explicit 3D modelling. The first row represents reenactment results, since the frontal image was used for initialization of both methods. We can see that our model does a much better job of modelling the face shape and the hair.}
    \label{fig:avatarsdk_comp}
\end{figure}

\begin{figure}[t]
    \centering    
    \setlength{\wid}{0.155\textwidth}
    \begin{tabular}{cccccc}
        \includegraphics[align=c,width=\wid,bmargin=0.1cm]{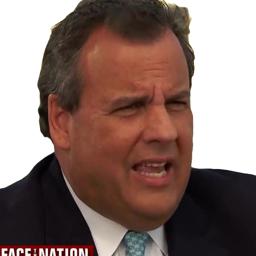} &
        \includegraphics[align=c,width=\wid,bmargin=0.1cm]{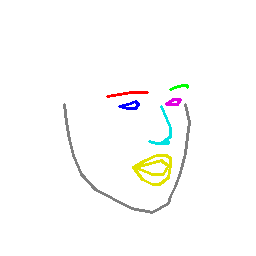} &
        \includegraphics[align=c,width=\wid,bmargin=0.1cm]{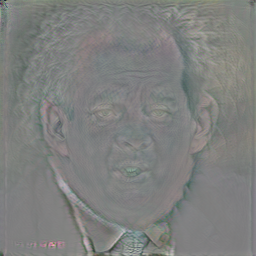} &
        \includegraphics[align=c,width=\wid,bmargin=0.1cm]{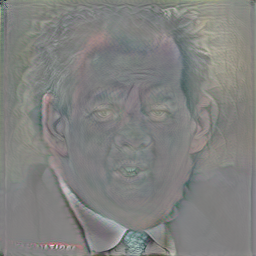} & 
        \includegraphics[align=c,width=\wid,bmargin=0.1cm]{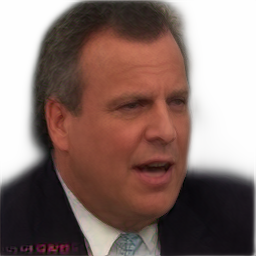} &
        \includegraphics[align=c,width=\wid,bmargin=0.1cm]{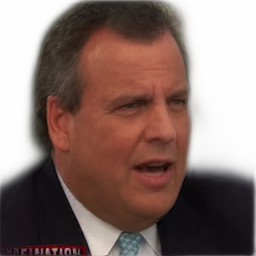} \\
        \includegraphics[align=c,width=\wid,bmargin=0.1cm]{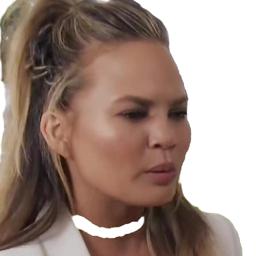} &
        \includegraphics[align=c,width=\wid,bmargin=0.1cm]{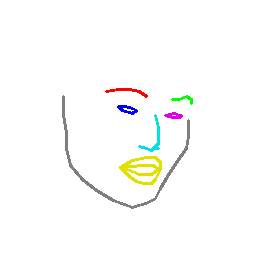} &
        \includegraphics[align=c,width=\wid,bmargin=0.1cm]{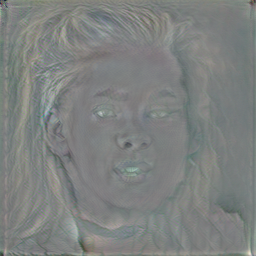} &
        \includegraphics[align=c,width=\wid,bmargin=0.1cm]{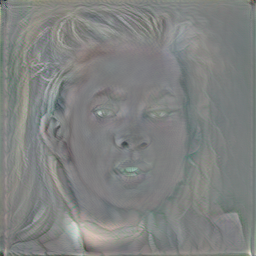} & 
        \includegraphics[align=c,width=\wid,bmargin=0.1cm]{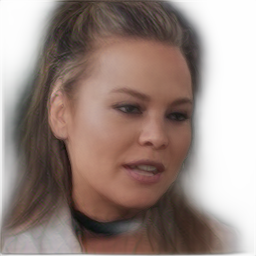} &
        \includegraphics[align=c,width=\wid,bmargin=0.1cm]{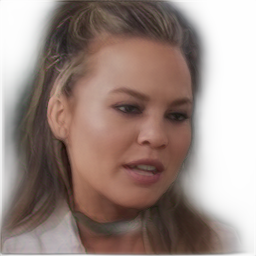} \\
        \includegraphics[align=c,width=\wid,bmargin=0.1cm]{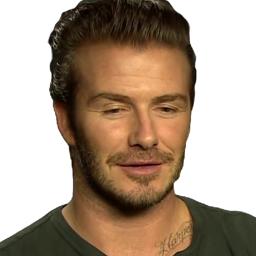} &
        \includegraphics[align=c,width=\wid,bmargin=0.1cm]{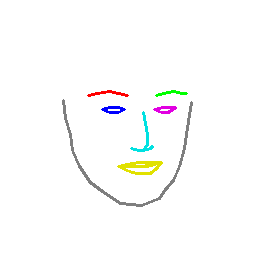} &
        \includegraphics[align=c,width=\wid,bmargin=0.1cm]{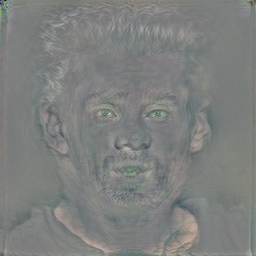} &
        \includegraphics[align=c,width=\wid,bmargin=0.1cm]{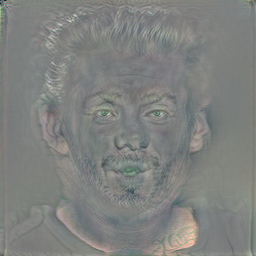} & 
        \includegraphics[align=c,width=\wid,bmargin=0.1cm]{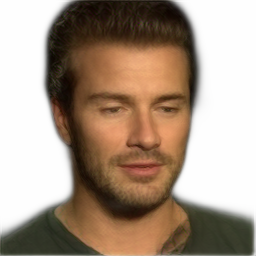} &
        \includegraphics[align=c,width=\wid,bmargin=0.1cm]{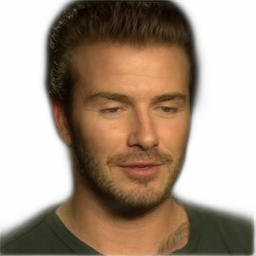} \\
        \includegraphics[align=c,width=\wid,bmargin=0.1cm]{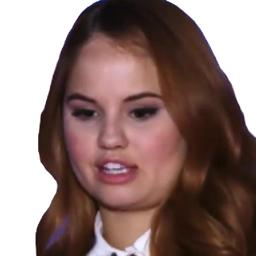} &
        \includegraphics[align=c,width=\wid,bmargin=0.1cm]{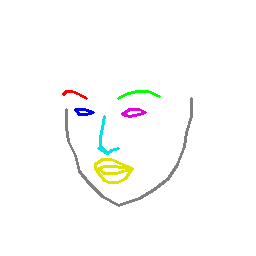} &
        \includegraphics[align=c,width=\wid,bmargin=0.1cm]{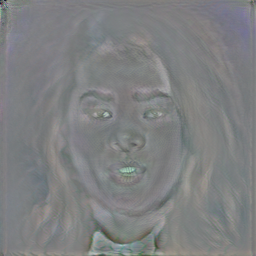} &
        \includegraphics[align=c,width=\wid,bmargin=0.1cm]{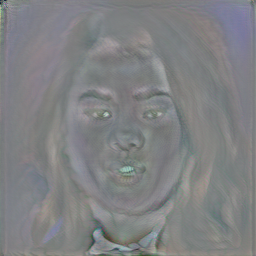} & 
        \includegraphics[align=c,width=\wid,bmargin=0.1cm]{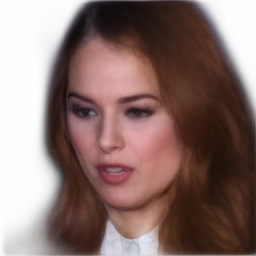} &
        \includegraphics[align=c,width=\wid,bmargin=0.1cm]{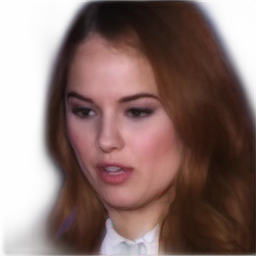} \\
        \textbf{Source} & \textbf{Pose} & \textbf{Texture} & \textbf{+ Upd.} & \textbf{Ours} & \textbf{+ Upd.}
    \end{tabular}
    \caption{Ablation study for the contribution of the texture updater on a VoxCeleb2-HQ dataset. The results are presented with and without the updater.}
    \label{fig:ext_ablation_study}
\end{figure}

\begin{figure}[t]
    \centering    
    \setlength{\wid}{0.155\textwidth}
    \begin{tabular}{cccc}
        \includegraphics[align=c,width=\wid,bmargin=0.1cm]{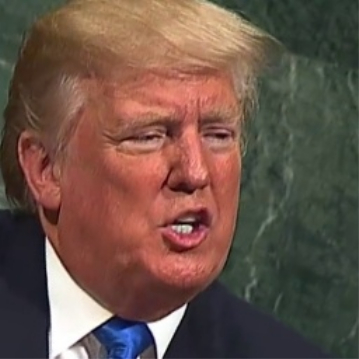} &
        \includegraphics[align=c,width=\wid,bmargin=0.1cm]{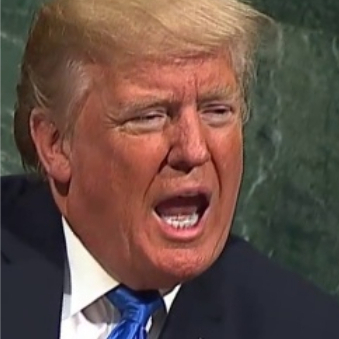} &
        \includegraphics[align=c,width=\wid,bmargin=0.1cm]{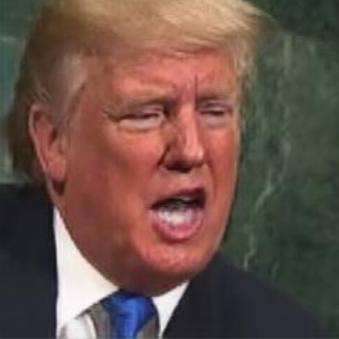} &
        \includegraphics[align=c,width=\wid,bmargin=0.1cm]{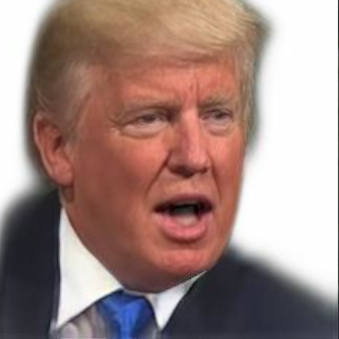} \\
        \includegraphics[align=c,width=\wid,bmargin=0.1cm]{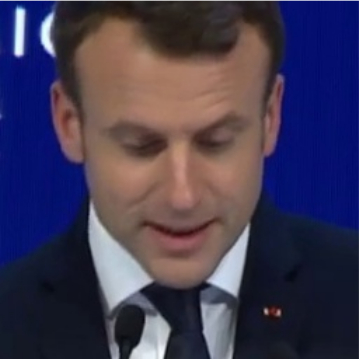} &
        \includegraphics[align=c,width=\wid,bmargin=0.1cm]{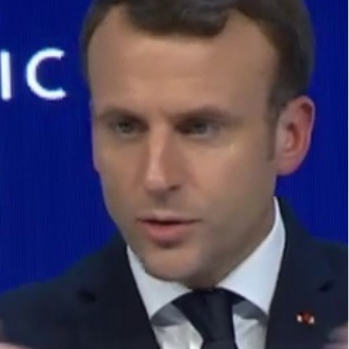} &
        \includegraphics[align=c,width=\wid,bmargin=0.1cm]{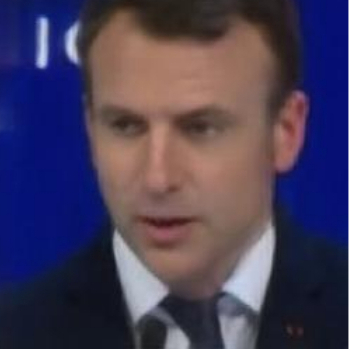} &
        \includegraphics[align=c,width=\wid,bmargin=0.1cm]{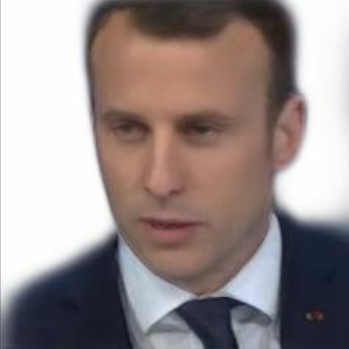} \\
        \includegraphics[align=c,width=\wid,bmargin=0.1cm]{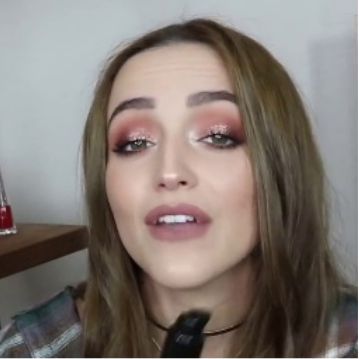} &
        \includegraphics[align=c,width=\wid,bmargin=0.1cm]{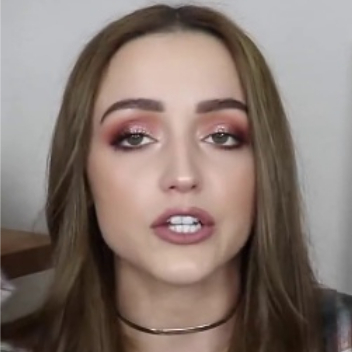} &
        \includegraphics[align=c,width=\wid,bmargin=0.1cm]{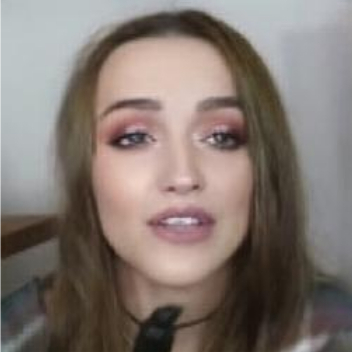} &
        \includegraphics[align=c,width=\wid,bmargin=0.1cm]{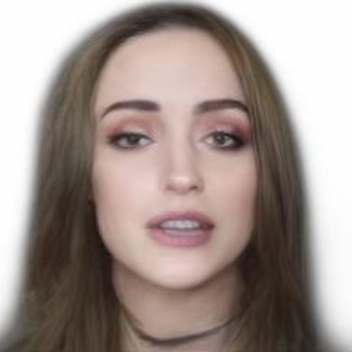} \\
        \includegraphics[align=c,width=\wid,bmargin=0.1cm]{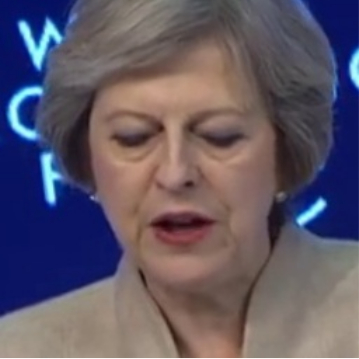} &
        \includegraphics[align=c,width=\wid,bmargin=0.1cm]{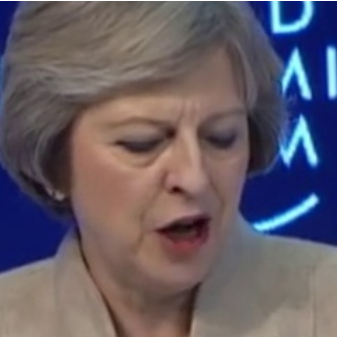} &
        \includegraphics[align=c,width=\wid,bmargin=0.1cm]{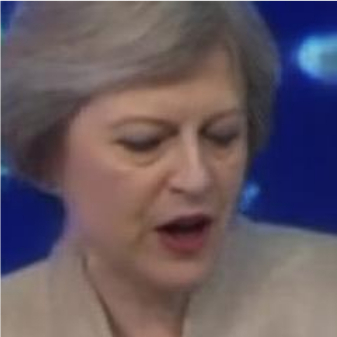} &
        \includegraphics[align=c,width=\wid,bmargin=0.1cm]{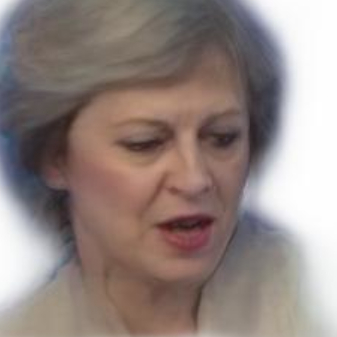} \\
        \textbf{Source} & \textbf{Target} & \textbf{\cite{Ha19}} & \textbf{Ours}
    \end{tabular}
    \caption{A comparison with MarioNETte~\cite{Ha19} system in a one-shot self-reenactment task. The results for~\cite{Ha19} are taken from the respective paper, as no source code is available. The evaluation of the computational complexity of this system was also beyond our reach since it would require re-implementation from scratch. However, since it utilizes an encoder-decoder architecture with a large number of channels~\cite{Ha19}, it can be assumed to have a similar complexity to the largest variant of FOMM~\cite{Siarohin19}. For our method, we use a medium-sized model. Lastly, the evaluation for~\cite{Ha19} is done on the same videos as training (on the hold-out frames), while our method is applied without any fine-tuning.}
    \label{fig:marionette_comp}
\end{figure}

\begin{figure}[t]
    \centering    
    \setlength{\wid}{0.148\textwidth}
    \begin{tabular}{cccccc}
        &
        \includegraphics[align=c,width=\wid,bmargin=0.1cm]{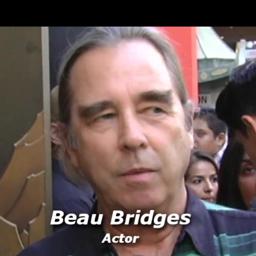} &
        \includegraphics[align=c,width=\wid,bmargin=0.1cm]{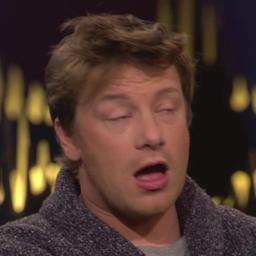} &
        \includegraphics[align=c,width=\wid,bmargin=0.1cm]{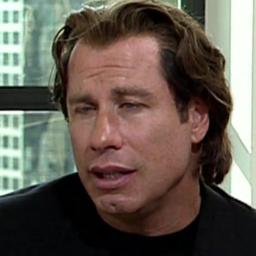} &
        \includegraphics[align=c,width=\wid,bmargin=0.1cm]{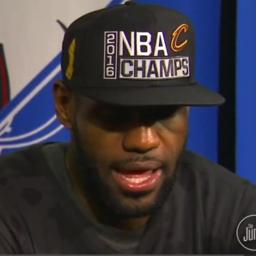} &
        \includegraphics[align=c,width=\wid,bmargin=0.1cm]{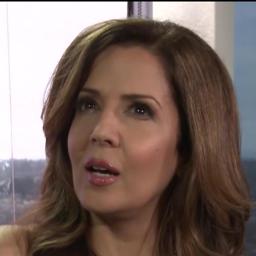} \\
        \includegraphics[align=c,width=\wid,bmargin=0.1cm]{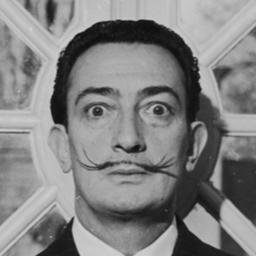} &
        \includegraphics[align=c,width=\wid,bmargin=0.1cm]{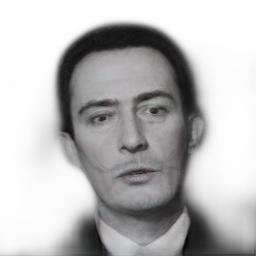} &
        \includegraphics[align=c,width=\wid,bmargin=0.1cm]{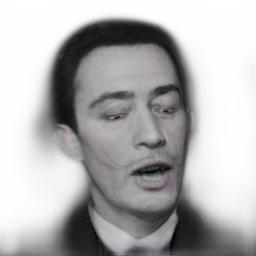} &
        \includegraphics[align=c,width=\wid,bmargin=0.1cm]{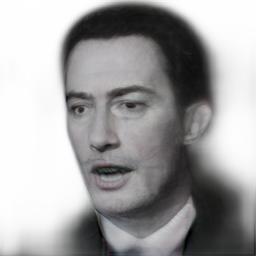} &
        \includegraphics[align=c,width=\wid,bmargin=0.1cm]{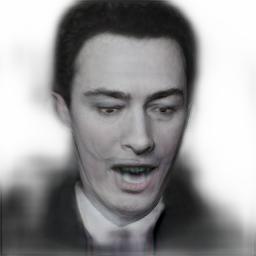} &
        \includegraphics[align=c,width=\wid,bmargin=0.1cm]{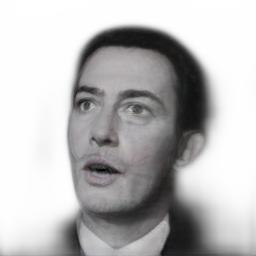} \\
        \includegraphics[align=c,width=\wid,bmargin=0.1cm]{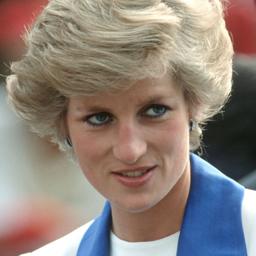} &
        \includegraphics[align=c,width=\wid,bmargin=0.1cm]{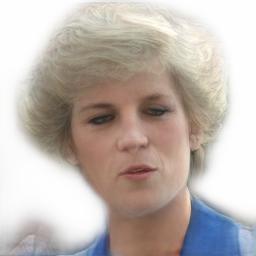} &
        \includegraphics[align=c,width=\wid,bmargin=0.1cm]{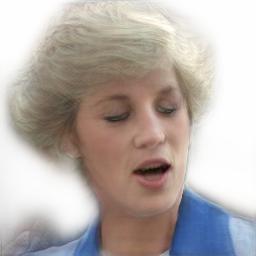} &
        \includegraphics[align=c,width=\wid,bmargin=0.1cm]{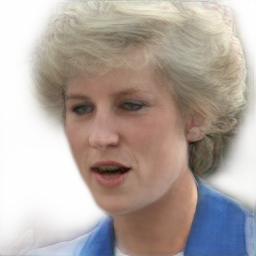} &
        \includegraphics[align=c,width=\wid,bmargin=0.1cm]{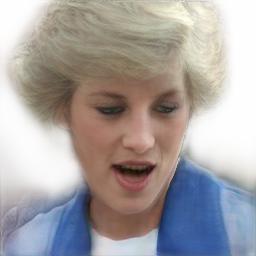} &
        \includegraphics[align=c,width=\wid,bmargin=0.1cm]{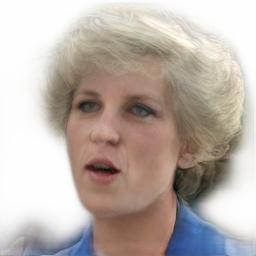} \\
        \includegraphics[align=c,width=\wid,bmargin=0.1cm]{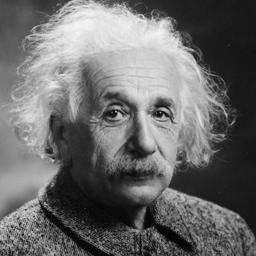} &
        \includegraphics[align=c,width=\wid,bmargin=0.1cm]{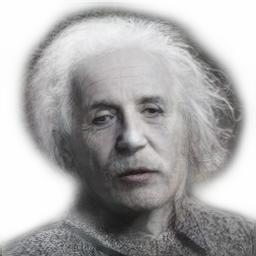} &
        \includegraphics[align=c,width=\wid,bmargin=0.1cm]{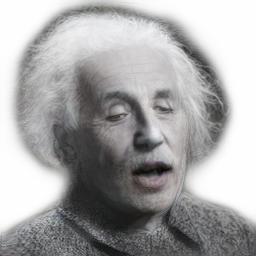} &
        \includegraphics[align=c,width=\wid,bmargin=0.1cm]{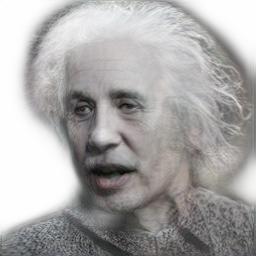} &
        \includegraphics[align=c,width=\wid,bmargin=0.1cm]{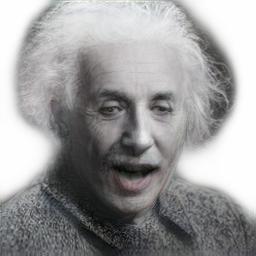} &
        \includegraphics[align=c,width=\wid,bmargin=0.1cm]{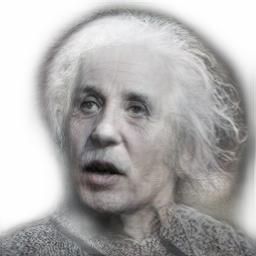} \\
        \includegraphics[align=c,width=\wid,bmargin=0.1cm]{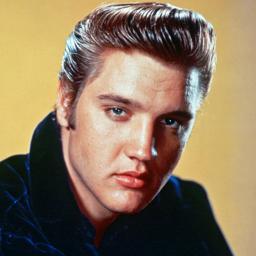} &
        \includegraphics[align=c,width=\wid,bmargin=0.1cm]{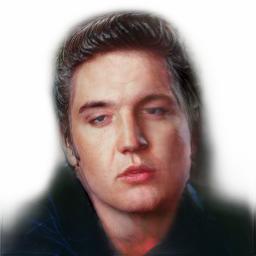} &
        \includegraphics[align=c,width=\wid,bmargin=0.1cm]{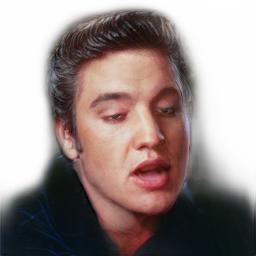} &
        \includegraphics[align=c,width=\wid,bmargin=0.1cm]{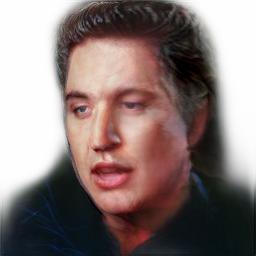} &
        \includegraphics[align=c,width=\wid,bmargin=0.1cm]{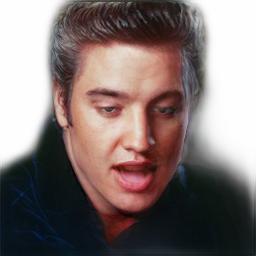} &
        \includegraphics[align=c,width=\wid,bmargin=0.1cm]{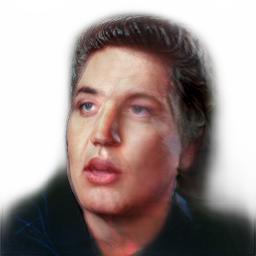} \\
        \includegraphics[align=c,width=\wid,bmargin=0.1cm]{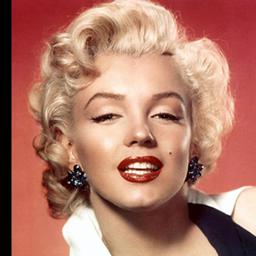} &
        \includegraphics[align=c,width=\wid,bmargin=0.1cm]{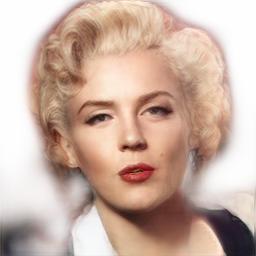} &
        \includegraphics[align=c,width=\wid,bmargin=0.1cm]{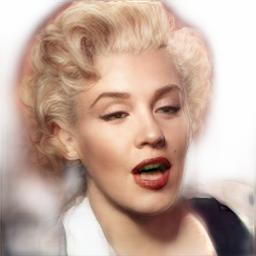} &
        \includegraphics[align=c,width=\wid,bmargin=0.1cm]{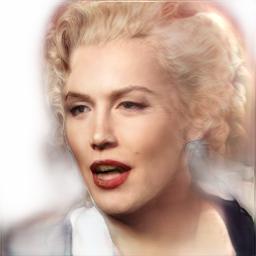} &
        \includegraphics[align=c,width=\wid,bmargin=0.1cm]{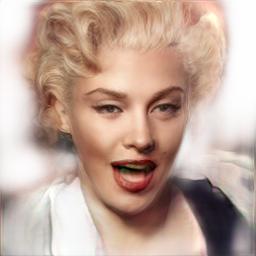} &
        \includegraphics[align=c,width=\wid,bmargin=0.1cm]{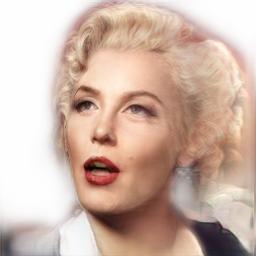} \\
        \includegraphics[align=c,width=\wid,bmargin=0.1cm]{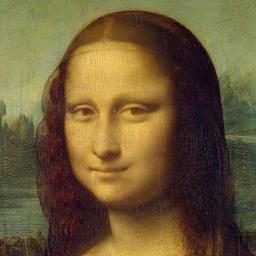} &
        \includegraphics[align=c,width=\wid,bmargin=0.1cm]{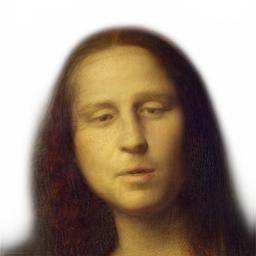} &
        \includegraphics[align=c,width=\wid,bmargin=0.1cm]{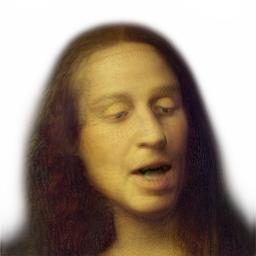} &
        \includegraphics[align=c,width=\wid,bmargin=0.1cm]{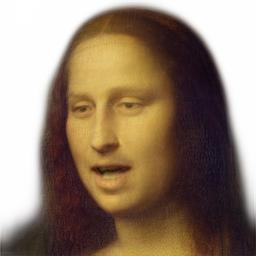} &
        \includegraphics[align=c,width=\wid,bmargin=0.1cm]{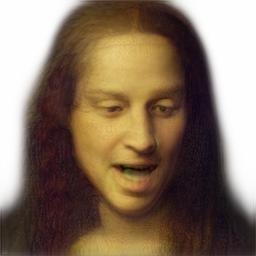} &
        \includegraphics[align=c,width=\wid,bmargin=0.1cm]{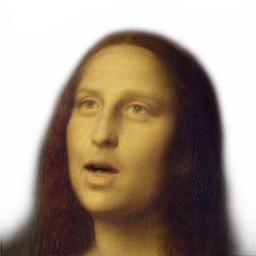} \\
        \includegraphics[align=c,width=\wid,bmargin=0.1cm]{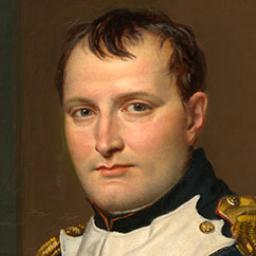} &
        \includegraphics[align=c,width=\wid,bmargin=0.1cm]{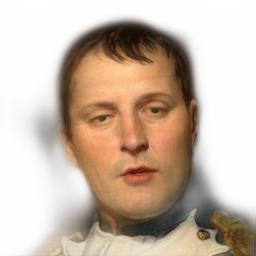} &
        \includegraphics[align=c,width=\wid,bmargin=0.1cm]{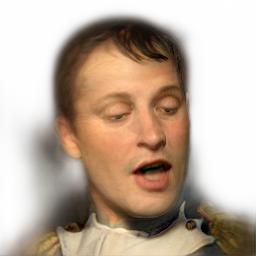} &
        \includegraphics[align=c,width=\wid,bmargin=0.1cm]{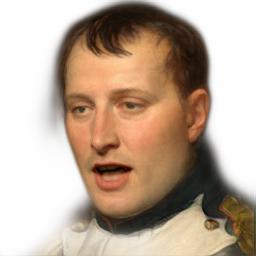} &
        \includegraphics[align=c,width=\wid,bmargin=0.1cm]{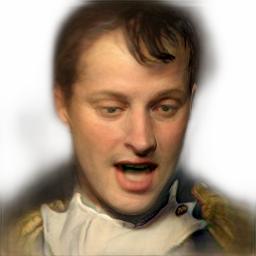} &
        \includegraphics[align=c,width=\wid,bmargin=0.1cm]{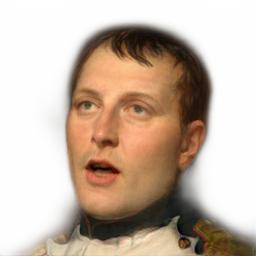} \\
        \includegraphics[align=c,width=\wid,bmargin=0.1cm]{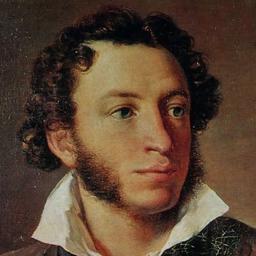} &
        \includegraphics[align=c,width=\wid,bmargin=0.1cm]{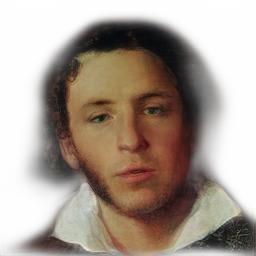} &
        \includegraphics[align=c,width=\wid,bmargin=0.1cm]{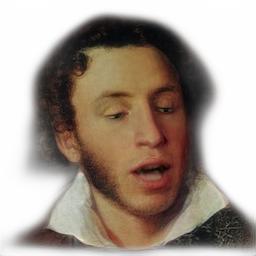} &
        \includegraphics[align=c,width=\wid,bmargin=0.1cm]{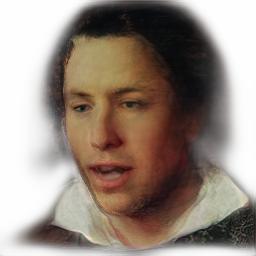} &
        \includegraphics[align=c,width=\wid,bmargin=0.1cm]{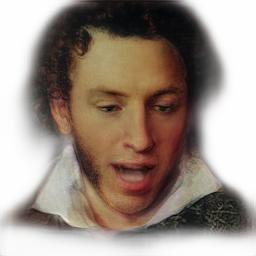} &
        \includegraphics[align=c,width=\wid,bmargin=0.1cm]{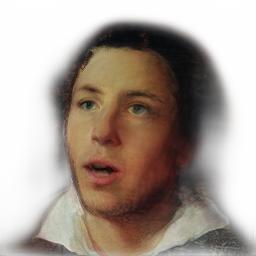} \\
        \textbf{Source} & \textbf{Driver 1} & \textbf{Driver 2} & \textbf{Driver 3} & \textbf{Driver 4} & \textbf{Driver 5}
    \end{tabular}
    \caption{The results for cross-person reenactment. While our method does preserve the texture of the original image, the driving identity leakage remains noticeable.}
    \label{fig:living_portraits}
\end{figure}

We provide extended quantitative data for our experiments in Table~\ref{tab:quant_comp_raw}, and additional qualitative comparisons in Figures~\ref{fig:ext_vc2_comp}-\ref{fig:vc2_family_comp}, which extend the comparisons provided in the main paper. We additionally perform a small comparison with a representative mesh-based avatar system~\cite{AvatarSDK} in Figure~\ref{fig:avatarsdk_comp} and compare our method with MarioNETte system~\cite{Ha19} in Figure~\ref{fig:marionette_comp}. Also we extend our ablation study to highlight the contribution of the texture enhancement network in the Figure~\ref{fig:ext_ablation_study}. Finally, we show cross-person reenactment results in Figure~\ref{fig:living_portraits}.

\begin{table}[!h]
    \centering
    \begin{tabular}{ l c c c c c c c}
         Method \ & \ LPIPS$\downarrow$ \ & \ SSIM$\uparrow$ \ & \ CSIM$\uparrow$ \ & \ NME$\downarrow$ \ & \ GMACs$\downarrow$ \ & Init. (ms)$\downarrow$ \ & Inf. (ms)$\downarrow$ \\
        \hline
        \multicolumn{8}{c}{Small models} \\
        \hline
        F-s V2V & 0.389 & - & 0.600 & 0.581 & 10.2 & - & - \\
        FOMM & \textbf{0.325} & - & \textbf{0.622} & 0.503 & 3.78 & - & - \\
        Ours & 0.392 & - & 0.540 & \textbf{0.475} & \textbf{1.08} & - & - \\
        \hline
        \multicolumn{8}{c}{Medium models} \\
        \hline
        F-s V2V & 0.368 & 0.419 & 0.604 & 0.461 & 18.2 & 4 & 22  \\
        FOMM & \textbf{0.311} & \textbf{0.553} & 0.638 & 0.478 & 13.9 & \textbf{3} & 13 \\
        Ours & 0.358 & 0.508 & \textbf{0.653} & \textbf{0.433} & \textbf{4.32} & 53 & \textbf{4} \\
        \hline
        \multicolumn{8}{c}{Large models} \\
        \hline
        NTH & 0.386 & - & 0.419 & 0.459 & 52.8 & - & - \\
        F-s V2V & 0.364 & - & 0.623 & 0.441 & 22.2 & - & - \\
        FOMM & \textbf{0.298} & - & \textbf{0.661} & 0.450 & 53.7 & - & - \\
        Ours & 0.356 & - & 0.655 & \textbf{0.428} & \textbf{17.3} & - & - \\
    \end{tabular}
    \caption{We present numerical data for the comparison of the models. Some of it duplicates the data available in Figure 5 of the main paper. F-s V2V denotes Few-shot Vid-to-Vid~\cite{Wang19}, FOMM denotes First Order Motion Model~\cite{Siarohin19}, and NTH denotes Neural Talking Heads~\cite{Zakharov19}. Here we also include SSIM evaluation, which we found to correlate with LPIPS, and therefore excluded it from the main paper. We also provide evaluation for initialization and inference time (in milliseconds) for the medium-sized models of each method, measured on NVIDIA P40 GPU. We did not include this measurement in the main paper since we cannot calculate it using target low-performance devices (due to difficulties with porting the competitor models to the SNPE~\cite{SNPE} framework), while evaluation on much more powerful (in terms of FLOPs) desktop GPUs may be an inaccurate way to measure the performance on less capable devices. We, therefore, decided to stick with MACs as our performance metric, which is more common in the literature, but still provide our obtained numbers for desktop GPUs here. We report median values out of a thousand iterations with random inputs.}\label{tab:quant_comp_raw}
\end{table}

\end{document}